\documentclass[journal]{IEEEtran}
\usepackage{amsmath,amsfonts}
\usepackage{algorithmic}
\usepackage{algorithm}
\usepackage{array}
\usepackage[caption=false,font=normalsize,labelfont=sf,textfont=sf]{subfig}
\usepackage{textcomp}
\usepackage{stfloats}
\usepackage{url}
\usepackage{verbatim}
\usepackage{graphicx}
\usepackage{cite}

\usepackage{bm}
\usepackage{multirow}
\usepackage{hyperref}

\usepackage[table]{xcolor}

\definecolor{lightgray}{gray}{0.9}

\usepackage{fancyhdr}
\begin{document}

\title{Simple Yet Effective Selective Imputation \\ for Incomplete Multi-view Clustering}

\author{
    Cai~Xu,
    Jinlong~Liu,
    Yilin~Zhang,
    Ziyu~Guan,
    Wei~Zhao,
    and~Xiaofei~He, \textit{Fellow, IAPR}

\thanks{Cai~Xu, Jinlong~Liu, Yilin~Zhang, Ziyu~Guan, and Wei~Zhao are with the School of Computer Science and Technology, 
    Xidian University, Xi'an, Shaanxi 710071, China.
    \textit{E-mail:\{cxu@, ljl\_liu@stu., ylzhang\_3@stu.,zyguan@,ywzhao@mail.\}xidian.edu.cn}.}

\thanks{Xiaofei~He is with the State Key Laboratory of CAD\&CG, Zhejiang
University, Hangzhou 310030, China. E-mail: 
    \textit{E-mail:xiaofeihe@cad.zju.edu.cn}.}
\thanks{(Corresponding author: Ziyu~Guan.)}

}



\maketitle

\begin{abstract}
Incomplete Multi-view Clustering (IMC) has emerged as a significant challenge in multi-view learning. A predominant line for IMC is data imputation; however, indiscriminate imputation can result in unreliable content. Recently, researchers have proposed selective imputation methods that use a post-imputation assessment strategy: (1) impute all or some missing values, and (2) evaluate their quality through clustering tasks. We observe that this strategy incurs substantial computational complexity and is heavily dependent on the performance of the clustering model. To address these challenges, we first introduce the concept of pre-imputation assessment. We propose an Implicit Informativeness-based Selective Imputation (SI$^3$) method for incomplete multi-view clustering, which explicitly addresses the trade-off between imputation utility and imputation risk. SI$^3$ evaluates the imputation-relevant informativeness of each missing position in a training-free manner, and selectively imputes data only when sufficient informative support is available. Under a multi-view generative assumption, SI$^3$ further integrates selective imputation into a variational inference framework, enabling uncertainty-aware imputation at the latent distribution level and robust multi-view fusion. Compared with existing selective imputation strategies, SI$^3$ is lightweight, data-driven, and model-agnostic, and can be seamlessly incorporated into existing incomplete multi-view clustering frameworks as a plug-in strategy. Extensive experiments on multiple benchmark datasets demonstrate that SI$^3$ consistently outperforms both imputation-based and imputation-free methods, particularly under challenging unbalanced missing scenarios. The code and datasets are available at:~\url{https://github.com/Starnever0/SI3}.
\end{abstract}


\section{Introduction}
\IEEEPARstart{M}{ulti-view} data has attracted increasing attention for its ability to provide more comprehensive descriptions of real-world entities or events. For example, on social media platforms, the user generated content may be composed of text view (e.g., descriptions or comments), image view (e.g., uploaded photos), and location view (e.g., GPS-based check-ins). These views collectively depict user behaviors: text reveals opinions or sentiments, images convey visual content, and location provides spatial context. By leveraging the consistency and complementarity across views, multi-view learning has achieved remarkable success in downstream tasks such as classification \cite{hanTrusted2021, seeland2021multi}, clustering \cite{11277376,11299478,luoSimple2024, moujahid2025advanced}, multimodal large language models \cite{yin2024survey} and embodied intelligence \cite{cangelosi2015embodied}. However, due to various practical limitations, real-world multi-view data are often incomplete, causing most existing methods to degrade or even fail inevitably. In this paper, we focus on the Incomplete Multi-view Clustering (IMC) problem, which integrates incomplete views to help identify essential grouping structure in an unsupervised manner.

\begin{figure}[t]
    \centering
    \includegraphics[width=0.475\textwidth]{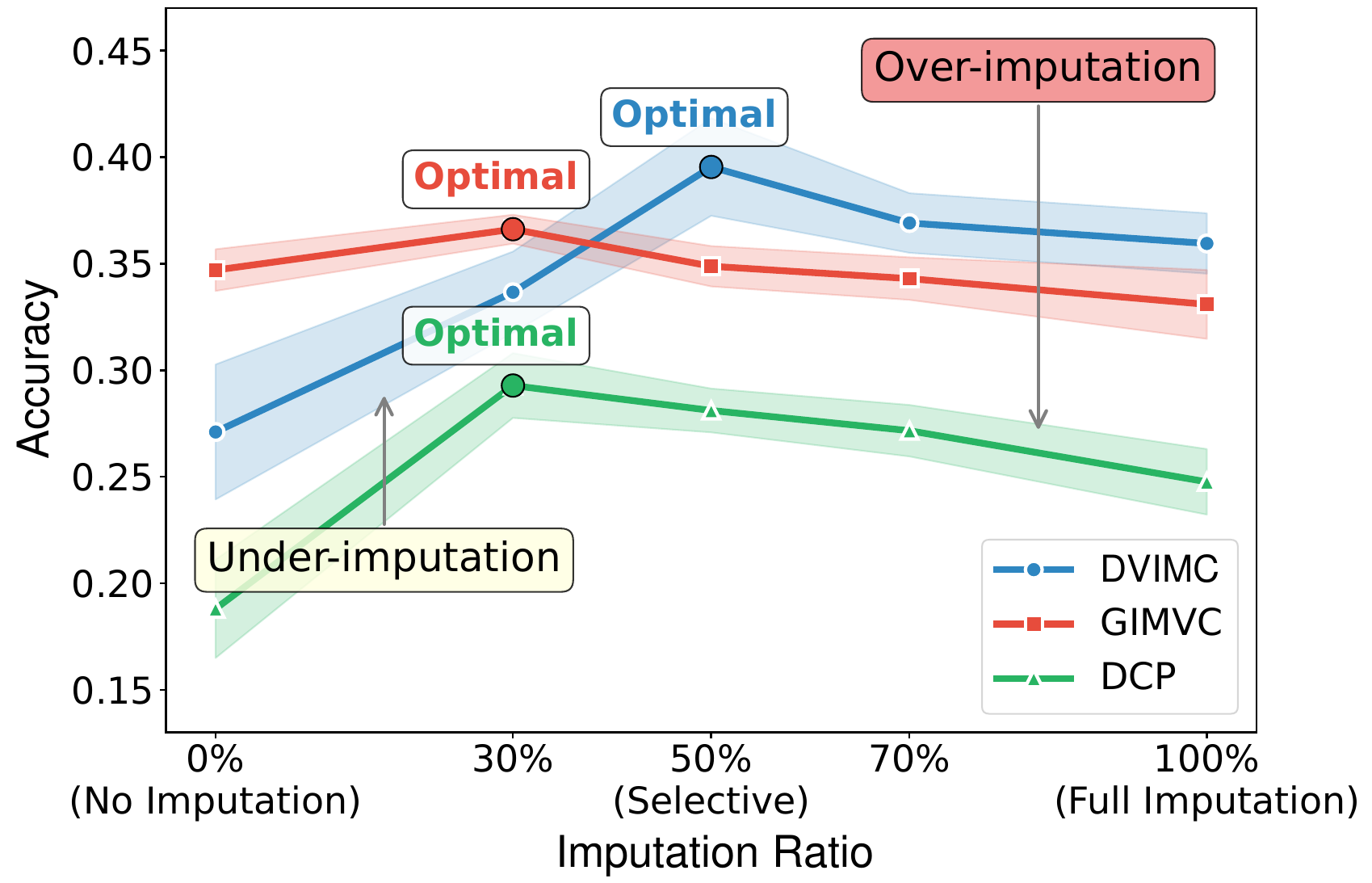}
    \caption{ 
    The clustering accuracies of three representative IMC methods enhanced by the proposed pre-imputation assessment strategy, which prioritizes imputing more informative missing positions in a training-free manner.}
    \label{fig:selection_ratio}
\end{figure}


A predominant line to solve this problem is imputation-based methods \cite{wang2018partial, Xu2019}, which estimate the missing positions via related positions (such as the positions of the same sample or cluster). While these methods enhance view completeness, their effectiveness highly depends on the reliability of the imputed data. However, indiscriminate imputation of all missing positions, without assessing the imputation-relevant informativeness provided by related observations related positions would involve unreliable content, especially under high missing rates \cite{yuanPrototype2025} or unbalanced view conditions \cite{fang2021unbalanced}. For instance, in user-generated content, the location view frequently exhibits higher missing rates due to privacy settings or device limitations, resulting in information scarcity when intra-cluster neighbors are also absent. In this condition, imputing such positions will introduce noisy information.




Recently, some researchers recognize this issue and propose selective imputation methods \cite{tangDeepSafeIncomplete2022, puAdaptive2024b} that utilize a post-imputation assessment strategy. These methods first impute all or partial positions, followed by clustering on the new data. They then evaluate the quality of the imputation through changes in loss or clustering consistency and reduce the influence of potentially noisy imputed positions by down-weighting or excluding them. This process is repeated multiple times until the model achieves optimal results. These methods have achieved good results, but this post-imputation assessment strategy still faces two new challenges: (1) the additional post-imputation assessment modules impose substantial computational complexity; (2) the post-imputation assessment strategy relies heavily on the clustering model to evaluate imputation quality, making the assessment sensitive to initialization, model stability, and intermediate training outcomes. When the clustering model is insufficiently trained, unstable, or exhibits poor performance, the imputation assessment becomes unreliable.

In the paper, we propose a simple yet effective selective imputation method (Implicit Informativeness-based Selective Imputation, SI$^3$) for this problem. The key motivations are: (1) not all positions should be imputed; (2) the decision to impute a position depends on the informativeness that can facilitate imputation for that position. Based on this, we introduce a pre-imputation assessment strategy, which uses a training-free manner to quantify the position's imputation-relevant informativeness from both intra-view and cross-view sources. In addition, to further mitigate the noise caused by unreliable imputations, we propose a distribution-level imputation strategy within a variational inference framework. The variance of the imputed distributions captures the uncertainty of imputations. This uncertainty-aware approach allows the multi-view fusion process to automatically down-weight unreliable views, thereby improving both robustness and semantic alignment across views. 

Extensive experiments demonstrate that our method consistently outperforms existing approaches across various missing rates. Fig.~\ref{fig:selection_ratio} shows an intuitive example: we enhance three representative IMC methods by the informativeness-based imputation position selection mechanism in our SI$^3$. The imputation ratio is gradually increasing from 0\% to 100\%. Experimental results indicate that both under-imputation and over-imputation yield poor outcomes: under-imputation causes information loss, while over-imputation introduces noise and degrades performance. Optimal performance is achieved at a moderate imputation ratio, indicating only partial imputation is beneficial. By identifying and imputing only the missing positions with high imputation-relevant informativeness, SI$^3$ effectively attains this favorable imputation regime and achieves superior performance. Our main contributions are summarized as follows:



\begin{itemize}
\item We have observed that selective imputation contributes significantly to addressing the important IMC problem (Fig.~\ref{fig:selection_ratio}). Moreover, we first introduce the concept of pre-imputation assessment to achieve a simple yet efficient selection of imputation positions.
\item The proposed pre-imputation assessment strategy operates at the data level, independent of specific models or clustering tasks. When integrated as a plug-in into representative baselines such as DCP\cite{linDual2022}, it incurs less than \textbf{2\%} additional runtime overhead while achieving up to \textbf{10.48\%} improvement in clustering accuracy, demonstrating strong robustness, adaptability, and practical efficiency across different baselines.
\item We design a distribution-level imputation strategy within a variational inference framework that models imputation uncertainty, implicitly down-weighting low-quality imputations during fusion to improve robustness.
\item Extensive experiments on diverse benchmark datasets demonstrate the effectiveness and generality of the proposed SI$^3$ framework. In particular, under severe and imbalanced missingness with an overall missing rate of 50\%, SI$^3$ achieves a \textbf{12.7\%} improvement in clustering accuracy over the imputation-free baseline DVIMC\cite{xu2024deep}.
\end{itemize}

\section{Related Works}

\subsection{Incomplete Multi-view Clustering.}

To tackle the challenges posed by incomplete multi-view data, existing IMC methods generally follow two main directions: (1) Imputation-based methods. Classical approaches leverage statistical correlations within and across views to estimate missing data~\cite{tran2017missing, liuEfficient2019}, or explicitly learning view-translation functions to recover the missing view from available ones \cite{linCOMPLETERIncompleteMultiView2021}. With the rise of deep generative models, a variety of methods have employed generative adversarial networks (GANs)~\cite{shang2017vigan, wang2018partial, zhang2020deep}, variational autoencoders (VAEs)~\cite{xuAudioVisual2024}, and more recently, diffusion models~\cite{wenDiffusionbased2024} to synthesize missing data. (2) imputation-free methods. In contrast, Imputation-free methods focus on designing missing-aware multi-view fusion or alignment strategies to directly learn the clustering structure from only the observed parts \cite{xu2022deep,baiGraphguided2024}. For example, ~\cite{tengURRLIMVC2024} introduces a unified representation learning framework based on an attention-driven autoencoder. \cite{xuAdaptive2023} avoids view imputation by learning a shared latent space through adaptive feature projection, which dynamically adjusts projection weights based on view quality and availability. 

While both paradigms have shown promise, they often suffer from performance degradation under severe or unbalanced missing conditions, as discussed in the Introduction. Our work departs from these extremes by proposing a lightweight, informativeness-based strategy that selectively imputes only reliable missing positions.


\subsection{Robust Utilization of Imputation.}
Except of exploring more accurate impute method, several studies adopt more cautious training schemes to alleviate negative effects from inaccurate imputation. One representative direction is the use of complete-case training paradigm: models are trained only on fully observed samples, while missing views are estimated at evaluation time using auxiliary prediction networks~\cite{linDual2022}. However, this method shares the limitations of imputation-free approaches, often yielding suboptimal performance in extreme scenarios. Another line of research employs more sophisticated techniques, involving explicitly evaluate imputation reliability after imputation by monitoring training dynamics and downstream performance \cite{yanDeep2025}. For instances, ~\cite{tangDeepSafeIncomplete2022} propose a bi-level optimization framework in which missing views are dynamically imputed from semantic neighbors, and the quality of these imputations is assessed through variations in the training loss. Based on this assessment, imputed samples that are deemed unreliable are adaptively down-weighted or excluded from subsequent optimization, thereby reducing the risk of performance degradation caused by inaccurate imputations. Similarly,~\cite{huangUnified2024a} use the half-quadratic minimization technique automatically weight different samples, alleviating the impact of outliers and unreliable restored data. More recently, ~\cite{hu2024reliable} introduces a reliable imputation guidance module that distinguishes intrinsic zeros from technical zeros using cluster-level confidence (based on zero rates, means, and variances), selectively imputing only the latter while jointly reconstructing sample- and feature-level structures. ~\cite{xi2025lrgr} proposes a two-stage quality-adaptive framework: a local refinement stage using cross-view contrastive learning and view-specific prototypes, followed by a global realignment stage with confidence-weighted pseudo-labels to mitigate distribution shifts from unreliable imputation. For downstream result, ~\cite{puAdaptive2024b} adopt a cluster-oriented latent-space imputation strategy, followed by a post-imputation evaluation of clustering outcomes. When the representations obtained after imputation yield inferior clustering results compared to those learned from unimputed data, the imputation process is discarded and subsequent training relies solely on the original representations. 

Despite their effectiveness, these methods predominantly adopt post-imputation assessment paradigms, where imputation quality is assessed only after missing positions have already been filled and incorporated into training. Such strategies either rely heavily on model-dependent training dynamics or introduce complex architectural constraints, resulting in limited granularity, scalability, and robustness. These limitations motivate us to propose a more general and lightweight pre-imputation assessment framework, which explicitly quantifies position-level imputation-relevant informativeness at the data level and selectively determines whether a missing position should be imputed before training, enabling safer and more principled utilization of imputed data.

\subsection{Variational Autoencoder}

Variational Autoencoders (VAEs)~\cite{kingma2013auto} have emerged as a generative framework with explicit uncertainty modeling, and they have been widely adopted in representation learning and clustering. By adopting diverse priors over latent representations and incorporating different generative assumptions, classical works have enhanced the clustering ability and discriminative power of VAEs, for example through Mixture-of-Gaussians (MoG) prior assumption~\cite{jiangVariational2017a,dilokthanakul2016deep}, graph-based constraints~\cite{yang2019deep}, or disentangled continuous and discrete factors~\cite{dupont2018learning}. 

In multi-view settings, the probabilistic nature of VAEs makes them particularly suitable for encoding different modalities into view-specific latent spaces and fusing them through reasonable principles. For example, suzuki et al.~\cite{suzukiJoint2016} introduced a joint multimodal VAE that enforces a shared latent representation across views, enabling coherent generation and cross-modal inference. He et al.~\cite{heM^2VAE2025} leverages type-specific latent variables and a Product-of-Experts (PoE) fusion to disentangle common and modality-specific representations. VariGANs~\cite{zhaoMultiView2018} combine variational inference with GANs, where VAEs capture cross-view global appearance and GANs refine details for high-quality multi-view image generation. Yin et al. \cite{yinShared2020} introduced adaptive weights together with a MoG prior to encourage clustering-friendly shared representations. Another important direction partitions the latent space into shared and private components, effectively disentangling cross-view commonality from view-specific complementary information~\cite{xu2021multi}.

For real-world scenarios with partially observed views, VAEs have been extended to incomplete multi-view learning and imputation. MultiVI~\cite{ashuach2023multivi} integrates partially missing multi-omics data by learning a unified latent space, enabling joint analysis despite incomplete modalities. CMVAE~\cite{caiRealize2024b} introduces a MoG latent representation that jointly generates all views, enabling posterior-based recovery of missing views and improving clustering and classification performance. Xu et al.~\cite{xu2024deep} address incomplete multi-view clustering by employing view-specific VAEs and aggregating their latent representations via a Product-of-Experts to obtain a robust shared representation. 

Despite these advances, most existing methods assume moderately missing or balanced scenarios and rely on direct aggregation or reconstruction from latent representations. Such strategies may become unstable under highly missing or unbalanced conditions, where imputation noise can propagate into the fused representation. Different from these approaches, our work performs imputation at the distribution level, operating directly on posterior parameters while explicitly modeling imputation uncertainty. This design enables more cautious and selective utilization of imputation, leading to more stable fusion and reliable latent representations for clustering.

\begin{figure*}[t] 
\centering 
\includegraphics[width=0.9\textwidth]{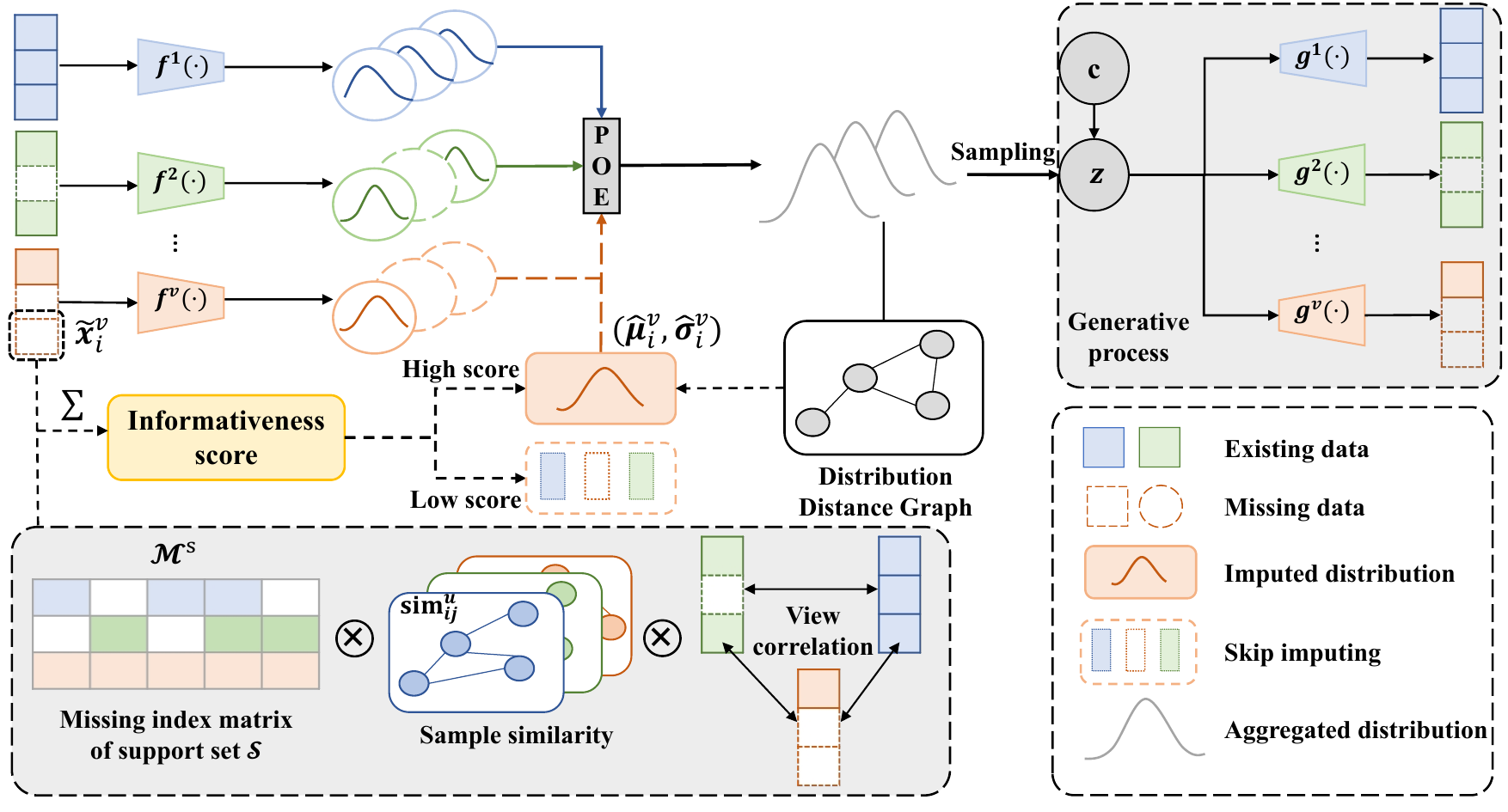} 
\caption{Overview of the SI$^3$ framework. An informativeness-based imputation position selection mechanism scores missing positions $\tilde{\bm{x}}_i^v$ based on intra-view similarity and cross-view correlation, imputing only those with sufficient support. Variational inference uses Product-of-Experts to aggregate view posteriors, with distribution-level imputation of missing views via latent neighbors. In the generative process, each observed view is generated from the cluster-aware latent representation $\bm{z}$. } 
\label{Fig.framework} 
\end{figure*}

\section{Method}
In this section, we present the Implicit Informativeness-based Selective Imputation (SI$^3$) method in detail, together with its implementation.

\subsection{Notations and Problem Statement}

Let $\{\{\bm{x}_i^v\}_{v=1}^V\}_{i=1}^N$ represent an incomplete multi-view dataset consisting of $N$ samples and $V$ views. Each $\bm{x}_i^v \in \mathbb{R}^{d_v}$ denotes the feature of the $i$-th sample in view $v$. $d_v$ is the feature dimension. The ground-truth label of the $i$-th sample is denoted by $y_i \in \{1, \dots, K\}$, where $K$ is the number of classes. We define a binary indicator matrix $\mathbf{M} \in \{0,1\}^{N \times V}$ to describe the observed pattern, where $\mathbf{M}_{i,v} = 1$ if $\bm{x}_i^v$ is observed, and $\mathbf{M}_{i,v} = 0$ otherwise. For notational convenience, we denote a missing position as $\tilde{\bm{x}}_i^v$ if $\mathbf{M}_{i,v} = 0$. The set of observed views for the $i$-th sample is defined as $\mathcal{V}_i = \{ v \mid \mathbf{M}_{i,v} = 1 \}$ and the set of observed samples for view $v$ is defined as $\mathcal{X}^v = \{ \bm{x}_i^v \mid \mathbf{M}_{i,v} = 1 \}$. The goal of IMC is to cluster the $N$ incomplete samples into $K$ groups.

\subsection{ The SI$^3$ Framework} 


In this section, we first outline the generative assumptions that form the foundation of the SI$^3$. We then introduce the pre-imputation assessment strategy, which calculates an informativeness score to each missing position to determine whether imputation is feasible. Finally, we detail the imputation strategy in a variational inference framework. The overall architecture of the SI$^3$ is illustrated in Fig.~\ref{Fig.framework}.

\subsubsection{The Multi-view Generative Assumption} 


In this paper, we introduce a multi-view generative assumption, which follows the formulation of the Deep Multi-view Gaussian Mixture Model (DMGMM) and consist of a three-step generative process: first, a shared discrete cluster assignment $\bm{c} \in \{0, 1\}^K$ is drawn from a categorical distribution with prior probabilities $\bm{\pi} \in \mathbb{R}^K$, where each $\pi_k$ denotes the prior probability of the $k$-th cluster and satisfies $\sum_{k=1}^K \pi_k = 1$. Then, a continuous latent variable $\bm{z} \in \mathbb{R}^d$ is sampled from a Gaussian distribution conditioned on the cluster assignment $\bm{c}$. Finally, given $\bm{z}$, each view-specific observation is independently generated from its corresponding conditional distribution. Under this assumption, the joint probability of an incomplete multi-view sample can be written as:
\begin{equation}
\label{eq:DMGMM}
p(\{\bm{x}_i^v\}_{v=1}^V, \bm{z}_i, \bm{c}_i) = p_{\bm{\theta}}(\{\bm{x}_i^v\}_{v=1}^V \mid \bm{z}_i) p(\bm{z}_i \mid \bm{c}_i) p(\bm{c}_i),
\end{equation}
where the joint likelihood can be factorized under the assumption that views are conditionally independent given $\bm{z}$:
\begin{equation}
\label{eq:factorized}
p_{\bm{\theta}}(\{\bm{x}_i^v\}_{v=1}^V \mid \bm{z}_i) = \prod_{v=1}^V p_{\bm{\theta}_v}(\bm{x}_i^v \mid \bm{z}_i).
\end{equation}

Each conditional distribution $p_{\bm{\theta}_v}(\bm{x}_i^v \mid \bm{z}_i)$ is modeled according to the data type of $\bm{x}_i^v$: a multivariate Gaussian distribution for real-valued features, with distribution parameters $[\bm{\mu}_{x_i^v}, \bm{\sigma}_{x_i^v}^2]$; or a multivariate Bernoulli distribution for binary features, with distribution parameter $\bm{\mu}_{x_i^v}$. These parameters are computed by the decoder $g^v(\bm{z}_i)$ with the trainable parameters $\bm{\theta}_v$.

In the \textbf{generative process} described above, the cluster assignment $\bm{c}_i$ first determines the latent representation $\bm{z}_i$, which serves as the underlying variable that generates multi-view observations ${\bm x_i^v}$. During \textbf{inference}, the direction is reversed: we aim to infer $\bm{z}_i$ and $\bm{c}_i$ from the observed views ${\bm x_i^v}$. Specifically, we first infer a latent representation $\bm{z}_i$ by integrating information from all available views, and then estimate the cluster assignment $\bm{c}_i$ based on the $\bm{z}_i$. Hence, the learning objective focuses on obtaining accurate and stable latent representations rather than reconstructing all missing observations.
However, \textbf{incomplete views provide limited evidence, which leads to uncertain or biased estimates of posterior distribution of $\bm{z}_i$ and consequently degrades clustering performance}.

There are two intuitive strategies to handle missing data. One is to directly recover missing data from the latent representation $\bm{z}_i$. However, this imputation paradigm contributes little to learning a more discriminative and cluster-consistent $\bm{z}_i$, and thus brings limited benefit to our clustering objective. Another strategy is to ignore the missing views and only aggregate observed ones~\cite{xu2024deep}. Yet under high missing rates or severe imbalance, this approach suffers from drastic degradation, and in extreme cases degenerates into a single-view solution that still outputs over-confident but biased representations.

To alleviate this issue, we regard imputation as injecting auxiliary information from other samples that share similar structures or semantics. By supplementing the incomplete observations with trustworthy evidence, the model obtains more support to infer a stable and cluster-consistent latent representation, effectively reducing the uncertainty in posterior estimates. Importantly, such auxiliary information is used not to directly fill the missing data, but to stabilize the evolving distribution parameters and guide latent inference.

Nevertheless, not all missing positions should be indiscriminately imputed. Some incomplete samples may lack sufficiently similar counterparts, while heterogeneous missing patterns may render cross-sample information unreliable. In such cases, naive imputation risks introducing noisy pseudo-observations that mislead posterior inference.

To address this, we introduce two complementary components: pre-imputation assessment of missing positions and variational inference with selective imputation. The former identifies positions with abundant and trustworthy evidence, ensuring that only well-supported missing positions are imputed, thereby mitigating the negative impact of unreliable auxiliary information. The latter models the uncertainty inherent in the imputation process by aggregating information from local neighbors at the distributional level, which not only stabilizes the missing-view likelihood approximation but also facilitates the learning of more robust and cluster-consistent latent representations.

\subsubsection{Informativeness-Based Imputation Position Selection}

\begin{figure}[t] 
\centering 
\includegraphics[width=0.9\columnwidth]{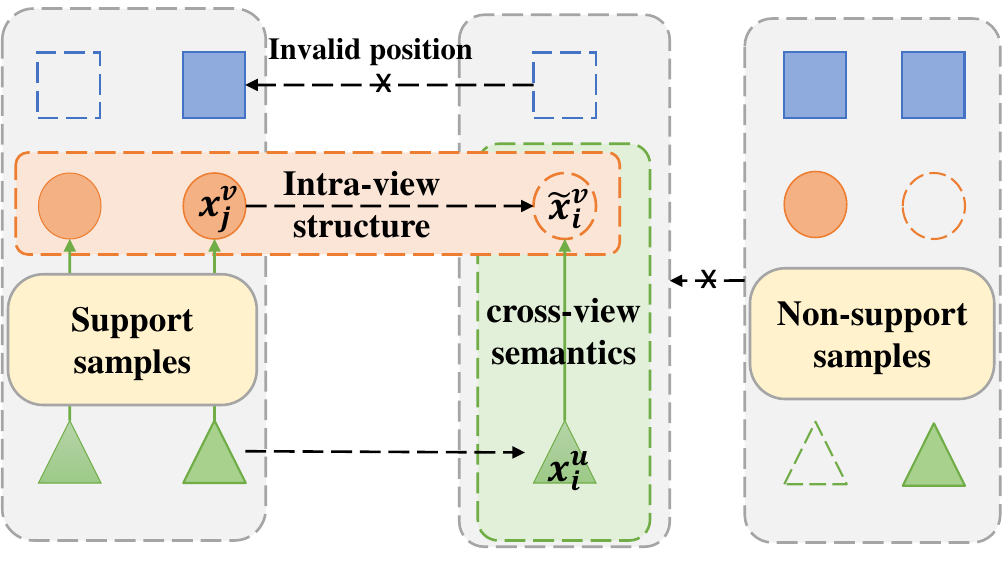} 
\caption{Explanation of the informativeness-based imputation position selection. When deciding whether to impute $\tilde{\bm{x}}_i^v$, we evaluate intra-view evidence (orange block) and cross-view evidence (green block). Non-support samples with varying observed patterns and invalid position of support samples are excluded.} 
\label{Fig.infoselect} 
\end{figure}

Analogous to how humans make educated guesses only when sufficient evidence is available, we argue that models should also perform imputation selectively. To this end, we introduce an informativeness-based imputation position selection mechanism for incomplete multi-view clustering. Unlike prior methods that impute all missing positions indiscriminately, our approach explicitly evaluates whether each missing position is sufficiently supported by two direct sources of evidence:

\begin{itemize}
    \item \textbf{Intra-view evidence.} As shown in the orange block of Fig.~\ref{Fig.infoselect}, discriminative neighbors in the same view $v$, i.e., those belonging to the same cluster or exhibiting high similarity, can provide support evidence for interpolation. Such neighbors preserve local manifold structures and category-consistent patterns, whereas using samples from other clusters may introduce contradictory signals and noise. Therefore, the availability of sufficiently many trustworthy neighbors within the same view is a key prerequisite for accurate imputation.
    
    \item \textbf{Cross-view evidence.} As illustrated in the green block of Fig.~\ref{Fig.infoselect}, the observed views of the same sample $\bm{x}_i$ primarily provide common and consistent semantics shared across views. Since different views describe the same underlying entity, their correlated features can reinforce the semantic consistency of imputation. Importantly, we emphasize leveraging this shared evidence rather than private, view-specific patterns, as the latter may introduce modality-specific bias and undermine clustering robustness. Thus, cross-view consistency serves as a reliable foundation for imputing missing values when intra-view support is limited.
\end{itemize}


Building upon these two complementary evidence sources, we construct a unified estimation framework that quantifies their information contribution via a position-specific support set. This allows us to compute an informativeness score $\mathrm{Info}(i,v)$ for each missing position and perform imputation only when sufficient support is available.

This quantification is inspired by the generative assumption, where samples with more similar latent posteriors are more likely to belong to the same Gaussian component (i.e., cluster). From this perspective, the contribution of different reference samples and views to imputing a missing position $\tilde{\bm{x}}_i^v$ can be naturally distinguished:
\begin{itemize}
    \item First, closer neighbors in latent space are expected to provide more reliable reference, thus receiving higher weights. 
    \item Second, samples with more observed views yield more accurate aggregated posteriors, making them stronger evidence in estimating distributional similarity. 
    \item Third, shared views between the target and its neighbors allow for more accurate computation of distributional distances, as they are conditioned on the same view-specific generative factors. 
    \item In particular, the current view $v$ (the view of missing position $(i,v)$) is of paramount importance, since it corresponds directly to the missing position and is generated under the same conditional distribution; therefore, its presence makes the similarity estimation especially trustworthy. Other views also contribute via the shared latent variable $\bm{z}_i$, but their influence critically depends on whether view $v$ is present: without it, we cannot determine whether their contribution reflects a consistent cluster membership or introduces noise.
\end{itemize}

We quantify the imputation-relevant informativeness of missing positions via the above principles. Each sample-view pair contributes differently based on its availability, similarity, and alignment with the target view. Specifically, the informativeness-based imputation position selection includes defining support samples, estimating information contributions, and computing informativeness scores. Next, each part will be introduced in detail.

\paragraph{Support Sample Definition}

In practice, not all samples are suitable for participating in the quantification of imputation-relevant informativeness for a missing position.
As illustrated on the right side of Fig.~\ref{Fig.infoselect}, samples that exhibit completely different missing patterns or lack any overlap in observed views should be excluded. This exclusion is not due to unreliability of the data, but rather because these samples do not provide a clear pathway for estimating the target view.

In most imputation paradigms, the estimation of a missing position relies on observable correspondences: either intra-view neighbors that share similar features, or cross-view relations established through jointly observed samples. When two samples share no common observed view, their potential statistical dependencies can only be implicitly captured during global network training, which is indirect, unstable, and not interpretable at the sample level. Therefore, we construct a set of support samples that serve as valid reference points for each missing position $(i,v)$.

Formally, we define a sample $\bm{x}_j$ as a support sample for imputing $\tilde{\bm{x}}_i^v$ if it satisfies two conditions:  
(i) $\bm{x}_j$ contains the target view $v$, ensuring direct reference in the view of interest; and  
(ii) $\bm{x}_j$ shares at least one observed view with $\bm{x}_i$, which enables meaningful estimation of similarity in latent space. Fig.~\ref{Fig.infoselect} illustrates how intra-view and cross-view signals are derived from support samples, and highlights the distinction between support and non-support samples. 

The complete set of support samples constitutes a position-specific support set $\mathcal{S}$\footnote{Note that both $\mathcal{S}$ and $\mathbf{M}^s$ are defined specifically with respect to the target missing position $(i,v)$. For brevity, we omit the subscripts $i,v$ when the context is unambiguous.}:
\begin{equation}
\mathcal{S} = \left\{ \bm{x}_j \mid \mathbf{M}_{j,v} = 1,~\mathcal{V}_i \cap \mathcal{V}_j \ne \emptyset \right\}.
\end{equation}

The cardinality $|\mathcal{S}|$ reflects the amount of potentially available structural context for the imputation. To indicate which parts of each support sample are useful for imputation, we introduce a binary mask matrix  $\mathbf{M}^s \in \{0,1\}^{|\mathcal{S}| \times V}$ associated with the support set, where $\mathbf{M}^s_{j,u} = 1$ indicates that view $u$ of support sample $\bm{x}_j$ is valid for contributing to the imputation. Non-target views are considered valid only if they overlap with the observed views of $\bm{x}_i$ (see Fig.~\ref{Fig.infoselect} for an invalid case).

\paragraph{Information Contribution Estimation}

To reflect the varying contributions of different sample positions to the calculation of imputation, we assign a continuous contribution score to each observed position $\bm{x}_j^v$ (intra-view) and $\bm{x}_j^u$ (cross-view), based on its estimated proximity to the target $\tilde{\bm{x}}_i^v$. Considering that closer samples and more correlated views provide more reliable references, we introduce two metrics:

1. \textbf{View correlation.} 
To quantify the consistency of shared semantics across views and provide reliable weighting for similarity approximation when the target view is missing, the inter-view correlation $\mathrm{corr}^{uv} \in (0,1]$ is computed in the latent space via Canonical Correlation Analysis (CCA)~\cite{andrewDeep2013}. 
Specifically, for views $u$ and $v$, we consider the $N_{uv} = |\mathcal{X}^u \cap \mathcal{X}^v|$ samples observed in both views and form their latent representations:
\begin{equation}
\mathbf{H}^v = [\bm{\mu}_1^v; \bm{\mu}_2^v; \dots; \bm{\mu}_{N_{uv}}^v] \in \mathbb{R}^{N_{uv} \times d_z}. 
\end{equation}

Given the latent matrices $\mathbf{H}^u$ and $\mathbf{H}^v$, CCA finds linear projections $\bm{a}^u, \bm{a}^v$ that maximize the correlation between projected features:
\begin{equation}
\mathrm{corr}^{uv} = \max_{\bm{a}^u, \bm{a}^v} \mathrm{corr}(\mathbf{H}^u \bm{a}^u, \mathbf{H}^v \bm{a}^v).
\end{equation}

To ensure that this correlation is computed by semantically meaningful latent representations, we first perform a lightweight pre-training using only the reconstruction loss during the initialization stage:
\begin{equation}
\label{eq:reco_loss}
    \mathcal{L}_{\text{rec}} = \sum_v |\bm{x}^v - g^v(\bm{\mu}^v)|_2^2,
\end{equation}
where $\bm{\mu}^v$ denotes the latent embedding output by the encoder $f^v(\bm{x}^v)$. At this stage, $\bm{\mu}^v$ is a deterministic representation without probabilistic semantics, and the pre-training is solely used to obtain a well-initialized latent space for reliable imputation-relevant informativeness estimation and efficient optimization. The same encoder–decoder architecture is later extended to a variational framework for uncertainty-aware modeling. Note that while using CCA, our framework is agnostic to the specific choice of correlation measure. For different models, better options such as mutual information or cross-view mapping loss could be preferred.

2. \textbf{Sample similarity}. To differentiate the contributions of intra-view neighbors and prioritize trustworthy samples more likely to belong to the same cluster or Gaussian component, intra-view similarity is defined as a normalized distance metric. Specifically, intra-view similarity $\mathrm{sim}_{ij}^u$ is computed as:
\begin{equation}
\mathrm{sim}_{ij}^u = \left( 1 - \frac{{\| \bm{x}_i^u - \bm{x}_j^u \|}_2}{\max_{k \ne l}
{\| \bm{x}_k^u - \bm{x}_l^u \|}_2} \right)^2,
\end{equation}
where the denominator is taken over all observed sample pairs in view $u$, ensuring scale-invariance and a range in $(0,1]$.

Since the target view $\tilde{\bm{x}}_i^v$ is unavailable, we approximate $\mathrm{sim}_{ij}^v$ by a weighted average over co-observed views:
\begin{equation}
\mathrm{sim}_{ij}^v = \sum_{u \in \mathcal{V}_i \cap \mathcal{V}_j} \mathrm{sim}_{ij}^u \cdot \frac{\mathrm{corr}^{uv}}{\sum_{u \in \mathcal{V}_i \cap \mathcal{V}_j} \mathrm{corr}^{uv}}.
\end{equation}

\paragraph{Informativeness Score Computation}
Using above metrics, we define the total contribution of intra-view evidence as:
\begin{equation}
\mathrm{Info}_{\text{intra}} = \sum_{j=1}^{|\mathcal{S}|} \mathrm{sim}_{ij}^v.
\end{equation}

For cross-view evidence $\bm{x}_i^u$ ($u \ne v$), the effectiveness of these evidence hinges on the quality of view-to-view semantic transfer, such as how well the mapping between views $u$ and $v$ has been learned. As this direct quality is challenging to measure, we approximate the contribution of $\bm{x}_i^u$ indirectly by aggregating Informativeness score from co-occurring support samples weighted by view correlation and local similarity:
\begin{equation}
\mathrm{Info}_{\text{cross}} =  \sum_{j =1}^{|\mathcal{S}|} \sum_{u\neq v} \mathrm{sim}_{ij}^u \cdot \mathrm{corr}^{uv} \cdot \mathbf{M}^s_{j,u}.
\end{equation}
This approximation captures how much the support sample $\bm{x}_j^u$ can help infer a reliable mapping from view $u$ to $v$, thus indirectly estimating the utility of $\bm{x}_i^u$ in filling $\tilde{\bm{x}}_i^v$. 

By setting $\mathrm{corr}^{vv} = 1$, the total Informativeness score for $\tilde{\bm{x}}_i^v$ is the sum of intra- and cross-view score:
\begin{equation}
\label{eq:info}
\begin{aligned}
    \mathrm{Info}(i,v) &= \mathrm{Info}_{\text{intra}} + \mathrm{Info}_{\text{cross}} \\
    &= \sum_{j =1}^{|\mathcal{S}|} \sum_{u=1}^{V} \mathrm{sim}_{ij}^u \cdot \mathrm{corr}^{uv} \cdot \mathbf{M}^s_{j,u}.
\end{aligned}
\end{equation}

We use this scalar value to selectively impute only those missing positions that are sufficiently supported. In detail, we assign the threshold $\tau$ connected with a proportion of missing data, and only perform imputation when $\mathrm{Info}(i,v)>\tau$.

\subsubsection{Variational Inference with Selective Imputation}
Our goal is to learn latent representations $\bm{z}_i$ and cluster assignments $\bm{c}_i$ for incomplete multi-view data, i.e., to infer the posterior distribution $p(\bm{z}_i, \bm{c}_i \mid \{\bm{x}_i^v\}_{v=1}^V)$. According to Bayes’ rule, the posterior distribution is generally intractable due to the complex integration over latent variables. To address this issue, we employ the mean-field variational approximation, which introduces an auxiliary distribution $q(\bm{z}_i, \bm{c}_i \mid \{\bm{x}_i^v\}_{v=1}^V)$ to approximate the intractable true posterior, which can be formulated as:
\begin{equation}
q(\bm{z}_i,\bm{c}_i \mid \{{\bm x}^v_i\})
= q(\bm{z}_i \mid \{{\bm x}_i^v\}) \,
  q(\bm{c}_i \mid \{{\bm x}_i^v\}).
\end{equation}
Following the mean-field factorization, we first focus on learning a shared latent representation $\bm{z}_i$, which serves as the basis for subsequent clustering.

\paragraph{Common Representation Learning}
For each observed view $v$, the approximate posterior is modeled by a Gaussian distribution:
\begin{equation}
q_{\bm{\phi}_v}(\bm{z}_i \mid \bm{x}_i^v) = \bm{\mathcal{N}}(\bm{z}_i \mid \bm{\mu}_i^v, (\bm{\sigma}_i^v)^2 \mathbf{I}),
\end{equation}
with $[\bm{\mu}_i^v, \bm{\sigma}_i^v] = f^v(\bm{x}_i^v)$, where $f^v$ denotes the pre-trained encoder in the initial stage and $ \mathbf{I}$ denotes the identity matrix. 

We adopt Product-of-Experts (PoE) for multi-view posterior aggregation due to its ability to emphasize agreement across views while naturally handling missing views:
\begin{equation}
\label{eq:PoE}
q(\bm{z}_i \mid \{\bm{x}_i^v\}_{v=1}^V) = \bm{\mathcal{N}}(\bm{z}_i \mid \bm{\mu}_i, (\bm{\sigma}_i)^2 \mathbf{I}),
\end{equation}
with
\begin{equation}
\label{eq:PoEpara}
\bm{\mu}_i = \frac{\sum_{v \in \mathcal{V}_i} \bm{\mu}_i^v / (\bm{\sigma}_i^v)^2}{\sum_{v \in \mathcal{V}_i} 1 / (\bm{\sigma}_i^v)^2}, \quad
(\bm{\sigma}_i)^2 = \frac{1}{\sum_{v \in \mathcal{V}_i} 1 / (\bm{\sigma}_i^v)^2},
\end{equation}
where \(\bm{\mu}_i\) and \((\bm{\sigma}_i)^2\) denote the aggregated mean and variance across available views \(\mathcal{V}_i\). This weights views by confidence (lower variance), but high missing rates or unbalanced view missing can destabilize the posterior. Missing views reduce terms in Eq.~\eqref{eq:PoEpara}, skewing $\bm{\mu}_i$ and underestimating $(\bm{\sigma}_i)^2$, especially when views are systematically absent, leading to biased estimates. To mitigate this, we impute missing view posteriors at the distribution level, ensuring a stable, reliable aggregated posterior that better captures the data distribution.

\paragraph{Distribution-level Selective Imputation}

For each missing position $(i,v)$ selected via Eq.~\eqref{eq:info} with $\mathrm{info}(i,v)>\tau$, we propose to impute its latent distribution parameters directly in the VAE latent space. Specifically, each sample–view pair is represented by a diagonal Gaussian posterior $q_i^v = \mathcal N(\bm{\mu}_i^v, \mathrm{diag}(\bm{\sigma}_i^v)^2).$

We first compute the parameters of aggregated posterior from available views using Eq.~\eqref{eq:PoEpara}, then identify its $k$ nearest neighbors $\mathcal{K}_{i,v}$ (with view $v$ present) based on the 2-Wasserstein distance between diagonal Gaussian posteriors:
\begin{equation}
\label{eq:distance}
W_2(q_i, q_j)
= \sqrt{|{\bm{\mu}}_i - {\bm{\mu}}_j|_2^2
+ |{\bm{\sigma}}_i - {\bm{\sigma}}_j|_2^2}.
\end{equation}

The latent parameters of the missing view are imputed via weighted averaging over these neighbors:
\begin{equation}
\label{eq:impute_mu}
\hat{\bm{\mu}}_i^{v} = \sum_{j \in \mathcal{K}_{i,v}} w_{ij} \cdot \bm{\mu}_j^{v},
\end{equation}
\begin{equation}
\label{eq:impute_sigma}
(\hat{\bm{\sigma}}_i^{v})^2 = \sum_{j \in \mathcal{K}_{i,v}} w_{ij} \cdot (\bm{\sigma}_j^{v})^2 + \text{Var}(\{\bm{\mu}_j^v : j \in \mathcal{K}_{i,v}\}),
\end{equation}
where $w_{ij} \propto \exp(-W_2(q_i, q_j))$, normalized via softmax. The additional variance term accounts for epistemic uncertainty during imputation.

Finally, the imputed posterior parameters are incorporated into Eq.~\eqref{eq:PoEpara} to re-estimate the aggregated latent representation $q(\bm{z}_i)$, which is then used for downstream clustering. This two-step aggregation—first across available views, then over selected neighbors—enables the model to leverage information from similar samples when data is scarce, resulting in a more stable and reliable latent distribution.

\paragraph{Shared Cluster Assignment}
Based on the stabilized latent posterior $q(\bm{z}_i)$, we further infer the shared cluster assignment. Specifically, a latent sample $\bm{z}_i^{(1)}$ is drawn via the reparameterization trick, and the posterior responsibilities are then computed accordingly:
\begin{equation}
\begin{aligned}
\label{eq:cluster}
&q(c_{ik} = 1 \mid \{\bm{x}_i^v\}_{v=1}^V) = p(c_{ik} = 1 \mid \bm{z}_i) \\
&=\frac{p(\bm{z}_i^{(1)} \mid c_{ik} = 1) p(c_{ik} =1)}{\sum_{\bm{c}} p(\bm{z}_i^{(1)} \mid \bm{c}_i) p(\bm{c}_i)},
\end{aligned}
\end{equation}
where $\bm{z}_i^{(1)} = \bm{\mu}_i + \bm{\sigma}_i \circ \bm{\epsilon}$ denotes one Monte Carlo sample of the latent variable obtained via the reparameterization trick and $\bm{\epsilon} \sim \bm{\mathcal{N}}(\bm{0}, \mathbf{I})$, $\circ$ denotes element-wise multiplication.

After identifying feasible positions for imputation, the variational inference framework is employed to explicitly model the uncertainty associated with the imputation process. This uncertainty-aware mechanism adaptively modulates the contributions of different views during fusion, thereby suppressing the impact of low-quality observations and unreliable imputations on the learned representations.

\subsection{Loss Function}
We adopt a variational training objective that combines the standard Evidence Lower Bound (ELBO)~\cite{kingma2013auto} with a cross-view coherence regularization to optimize the model. This formulation enables effective multi-view representation learning while handling missing views and preserving latent structure.

The ELBO component promotes data reconstruction and cluster-aware latent encoding, and is formulated as:
\begin{equation}
\begin{aligned}
&\bm{\mathcal{L}}_{\mathrm{ELBO}}(\{\bm{x}^{v}\}_{v=1}^{V})  \\
&= \mathbb{E}_{q_{\bm{\phi}}(\bm{z} \mid \{\bm{x}^{v}\}_{v=1}^V)} \left[ \sum_{v \in \mathcal{V}_i} \log p_{\bm{\theta}_v}(\bm{x}^{v} \mid \bm{z}) \right] \\
&- \mathbb{E}_{q_{\bm{\phi}}(\bm{c} \mid \{\bm{x}^v\}_{v=1}^V)} \left[ D_{\mathrm{KL}} \left(q_{\bm{\phi}}(\bm{z} \mid \{\bm{x}^{v}\}_{v=1}^V) \,\|\, p(\bm{z} \mid \bm{c})\right) \right] \\
&- D_{\mathrm{KL}} \left(q(\bm{c} \mid \{\bm{x}^{v}\}_{v=1}^V) \,\|\, p(\bm{c}) \right).
\end{aligned}
\end{equation}

To encourage consistency between the aggregated posterior and the view-specific posteriors, we include a coherence regularization term:
\begin{equation}
\begin{aligned}
&\bm{\mathcal{L}}_{\mathrm{CH}}(\{\bm{x}^v\}_{v=1}^V) \\
&= \sum_{v \in \mathcal{V}_i} -\frac{1}{|\mathcal{V}_i|} D_{\mathrm{KL}}\left(q_{\bm{\phi}}(\bm{z} \mid \{\bm{x}^v\}_{v=1}^V) \,\|\, q_{\bm{\phi}_v}(\bm{z} \mid \bm{x}^{v})\right).
\end{aligned}
\end{equation}

We optimize this objective using the reparameterization trick and stochastic gradient variational Bayes. The overall training loss is defined as:
\begin{equation}
\label{eq:loss}
\bm{\mathcal{L}} = \bm{\mathcal{L}}_{\mathrm{ELBO}} + \alpha \cdot \bm{\mathcal{L}}_{\mathrm{CH}},
\end{equation}
where \( \alpha \) is a hyperparameter balancing reconstruction and coherence.

To better understand the training process of our proposed framework, we detail the full optimization procedure in Algorithm~\ref{alg:algorithm}. 

\begin{algorithm}[t]
\begin{algorithmic}[1] 
\caption{Optimization of the proposed method}
\label{alg:algorithm}
\STATE \textbf{Input}: Incomplete multi-view dataset $\{\{\bm{x}_i^v\}_{v=1}^V\}_{i=1}^N$ with indicator matrix $\mathbf{M}$; Number of clusters $K$;Informativeness score threshold $\tau$; Regularization parameter $\alpha$.
\STATE Initialize network parameters $\{\bm{\theta}_v, \bm{\phi}_v\}_{v=1}^V$ and GMM parameters $\{\pi_k, \bm{\mu}_k, \bm{\sigma}_k^2\}_{k=1}^K$, get the initial latent representation $\mathbf{H}^v$.
\STATE Calculate $\mathrm{Info}(i,v)$ for each position $(i,v)$ using  Eq.~\eqref{eq:info}
\WHILE{not reaching the maximal epochs}
    \STATE Encode observed views to get $\{\bm{\mu}_i^v, \bm{\sigma}_i^v\}_{v\in \mathcal{V}_i}$.
    \STATE Aggregate original available posterior parameters to obtain $\{\bm{\mu_i}, (\bm{\sigma}_i)^2\}$ by Eq.~\eqref{eq:PoEpara}.
    \STATE Identify neighbors based on aggregated distribution.
    \FOR{each missing position $(i,v)$}
        \IF{$\mathrm{Info}(i,v) > \tau$}
            \STATE Impute latent parameters $\{\hat{\bm{\mu}}_i^v, \hat{\bm{\sigma}}_i^v\}$ via weighted averaging (Eq.~\eqref{eq:impute_mu} - Eq.~\eqref{eq:impute_sigma})
        \ENDIF
    \ENDFOR
    \STATE Re-aggregate posterior parameters including imputations (Eq.~\eqref{eq:PoEpara}).
    \STATE Compute $q(\bm{c} \mid \{\bm{x}^v\}_{v=1}^V)$ by Eq.~\eqref{eq:cluster}.
    \STATE Decode reconstructed views $\{p_{\theta_v}(\bm{x}^v \mid \bm{z})\}_{v\in \mathcal{V}_i}$ via view-specific decoders with reparameterization trick.
    \STATE Update all parameters by maximizing Eq.~\eqref{eq:loss}.
\ENDWHILE
\STATE Assign each sample $i$ to the cluster with maximum posterior probability
\end{algorithmic}
\end{algorithm}

\section{Experiments}

\subsection{Experimental Setup}

\paragraph{Datasets} 
We evaluate our method on four widely used real-world multi-view datasets: (1) Caltech7-5V~\cite{fei2004learning,li2015large} contains 1474 pictures of objects belonging to 7 classes and all images are described with five types of features (254 CENTRIST, 512 GIST, 1984 HOG, 928 LBP, and 40 wavelet moments). (2) Scene-15~\cite{fei2005bayesian} consists of 4,485 samples from 15 commonly seen indoor and outdoor scene categories, with each category containing 200–400 images features. It provides diverse scene samples covering natural environments (e.g., coast, forest), man-made environments (e.g., office, bedroom), and mixed scenes (e.g., street, store). (3) COIL100~\cite{nene1996columbia} includes 7,200 images from 100 object categories, and each object is captured from different angles. Three types of features are used as distinct views, namely Isometric Projection (ISO), Linear Discriminant Analysis (LDA), and Neighborhood Preserving Embedding (NPE). (4) Multi-Fashion~\cite{xiao2017fashion} is a multi-view extension of the Fashion-MNIST dataset, where different feature extractors are used to construct heterogeneous representations for each image. The details of these datasets are summarized in TABLE~\ref{table:dataset}.

\begin{table}[t]
\centering
\caption{Details of the datasets used in our experiments}
\setlength{\tabcolsep}{1.2mm}{
\resizebox{\columnwidth}{!}{
\begin{tabular}{cccc}
\hline
Dataset & Sample & Cluster & Dimensions\\
\hline
Caltech7-5V    & 1,474   & 7     & 40/254/1984/512/928 \\
Scene-15       & 4,485   & 15    & 20/59/40 \\
COIL100        & 7,200   & 100   & 30/99/30 \\
Multi-Fashion  & 10,000  & 10    & 784/784/784 \\
\hline
\end{tabular}
}}
\label{table:dataset}
\end{table}

\paragraph{Baselines} 
We compare our approach with seven IMC baselines, which can be categorized into three main types based on their strategies for handling missing data: imputation-based methods, imputation-free methods, and selective imputation methods:
\begin{itemize}
\item \textbf{Imputation based methods:} \textbf{PMIMC}~\cite{yuanPrototype2025}, the current state-of-the-art among imputation-based methods, performs prototype-based imputation by aligning incomplete samples with learned prototypes across views.
\textbf{CPSPAN}~\cite{jinDeep2023} employs a structure embedding imputation strategy to align features between views by filling in missing embeddings using nearby neighbors from other views.
\item \textbf{Imputation-free methods:} \textbf{GIMVC}~\cite{baiGraphguided2024} maximizes the utilization of existing information by fusing feature and graph information from observed samples.
\textbf{DIMVC}~\cite{xu2022deep} avoids both imputation and feature fusion, instead projecting multi-view representations into a shared high-dimensional space through nonlinear mappings.
\textbf{DVIMC}~\cite{xu2024deep}, often achieving competitive or superior performance among imputation-free approaches, aggregates representations from observed views using the Product-of-Experts (PoE) approach.
\item \textbf{Selective imputation methods:} \textbf{DCP}~\cite{linDual2022} performs contrastive learning only on complete samples and applies imputation before evaluation.
\textbf{DSIMVC}~\cite{tangDeepSafeIncomplete2022} learns a weighting function that down-weights unreliable imputed samples to mitigate the negative impact of inaccurate imputations.
\end{itemize}

\paragraph{Implementation Details}

To better reflect real-world conditions, this work investigates incomplete multi-view clustering under unbalanced and more challenging missing scenarios. Instead of setting a global ratio of incomplete samples, we simulate such situations by randomly removing views from each sample with different probabilities, and define the missing rate as the proportion of missing positions across all views, which is more standard and stricter than sample-level ratio. Clustering performance is assessed with Accuracy (ACC), Normalized Mutual Information (NMI), and Adjusted Rand Index (ARI), averaged over ten runs. For hyperparameters, we empirically set the regularization coefficient $\alpha$ to 5 for Caltech7-5V and COIL100, 10 for Multi-Fashion and Scene-15. All experiments run on an NVIDIA 5070Ti GPU.

\subsection{Experimental Evaluation}

To comprehensively evaluate the effectiveness, interpretability, and generality of the proposed Implicit Informativeness-based Selective Imputation (SI$^3$) method, we design experiments to answer the following key questions:
\begin{itemize}

    \item Q1: Overall Effectiveness. Does SI$^3$ achieve superior clustering performance compared to state-of-the-art imputation-based and imputation-free methods under various missing rates?
    \item Q2: Informativeness and Interpretability. How do Informativeness scores reflect the heterogeneity of missing positions, and how effectively do they guide the selective imputation process?
    \item Q3: Robustness and Imputation Ratio Sensitivity. How robust is SI$^3$ to different degrees of missingness and varying selective imputation ratios? Does the new imputation mechanism mitigate performance degradation under high missing rates?
    \item Q4: Generality and Plug-in Capability of IBSI. Can the proposed IBSI module be seamlessly integrated into other IMC frameworks to enhance their robustness and accuracy?
    \item Q5: Sensitivity to the Regularization Coefficient. How does the regularization parameter $\alpha$ affect the clustering performance of SI$^3$?
\end{itemize}

\paragraph{Q1: Overall Effectiveness}

\begin{table*}[t]
\vspace{0pt}
\centering
\caption{Clustering results of all methods on four datasets. The best results are highlighted in bold.}
\small
\setlength{\tabcolsep}{1.4mm}
\begin{tabular}{cc|ccc|ccc|ccc|ccc|ccc}
\hline
                             \multicolumn{2}{c|}{Missing rates} & \multicolumn{3}{c|}{0.1} & \multicolumn{3}{c|}{0.2} & \multicolumn{3}{c|}{0.3} & \multicolumn{3}{c|}{0.4} & \multicolumn{3}{c}{0.5}\\ \hline
                             \multicolumn{2}{c|}{Methods}       & \multicolumn{1}{c}{ACC} & \multicolumn{1}{c}{NMI} & ARI & \multicolumn{1}{c}{ACC} & \multicolumn{1}{c}{NMI} & ARI & \multicolumn{1}{c}{ACC} & \multicolumn{1}{c}{NMI} & ARI & \multicolumn{1}{c}{ACC} & \multicolumn{1}{c}{NMI} & ARI & \multicolumn{1}{c}{ACC} & \multicolumn{1}{c}{NMI} & ARI \\ \hline
\multirow{8}{*}{\rotatebox{90}{\textbf{Caltech7-5V}}}      & PMIMC   & 0.849 & 0.798 & 0.753 & 0.822 & 0.769 & 0.719 & 0.824 & 0.751 & 0.702 & 0.794 & 0.711 & 0.650 & 0.762 & 0.677 & 0.604 \\
                             & CPSPAN   & 0.784 & 0.729 & 0.660 & 0.776 & 0.691 & 0.632 & 0.763 & 0.665 & 0.601 & 0.657 & 0.613 & 0.516 & 0.650 & 0.600 & 0.500\\
                             & GIMVC  & 0.871 & \textbf{0.825} & 0.776 & 0.859 & 0.800 & 0.754 & 0.844 & 0.801 & 0.746 & 0.841 & 0.779 & 0.731 & 0.825 & \textbf{0.748} & 0.698\\
                             & DIMVC  & 0.813 & 0.701 & 0.685 & 0.837 & 0.726 & 0.710 & 0.833 & 0.719 & 0.697 & 0.717 & 0.663 & 0.497 & 0.554 & 0.452 & 0.353 \\
                             & DCP  & 0.628 & 0.642 & 0.538 & 0.570 & 0.577 & 0.425 & 0.484 & 0.554 & 0.350 & 0.469 & 0.458 & 0.282 & 0.403 & 0.309 & 0.099 \\
                             & DSIMVC   & 0.661 & 0.562 & 0.588 & 0.661 & 0.560 & 0.579 & 0.563 & 0.462 & 0.486 & 0.624 & 0.506 & 0.531 & 0.616 & 0.494 & 0.534 \\
                             & DVIMC    & 0.872 & 0.784 & 0.755  & 0.897 & 0.810 & 0.791  & 0.890 & 0.803 & 0.784  & 0.794 & 0.702 & 0.654 & 0.516 & 0.439 & 0.327  \\
                             \rowcolor{lightgray}
                             & \textbf{SI$^3$}   & \textbf{0.897} & 0.811 & \textbf{0.788} & \textbf{0.904} & \textbf{0.821} & \textbf{0.805} & \textbf{0.898} & \textbf{0.811} & \textbf{0.797} & \textbf{0.881} & \textbf{0.794} & \textbf{0.766} & \textbf{0.840} & 0.736 & \textbf{0.712} \\ \hline
\multirow{8}{*}{\rotatebox{90}{\textbf{Scene-15}}}   & PMIMC   & 0.405 & 0.413 & 0.234 & 0.401 & 0.422 & 0.236 & 0.404 & 0.427 & 0.241 & 0.394 & \textbf{0.407} & 0.240 & 0.325 & 0.327 & 0.172 \\
                             & CPSPAN   & 0.405 & 0.401 & 0.239 & 0.403 & 0.395 & 0.234 & 0.391 & 0.385 & 0.224 & 0.386 & 0.383 & 0.222 & 0.390 & \textbf{0.390} & 0.222\\
                             & GIMVC   & 0.402 & 0.423 & 0.244 & 0.401 & 0.406 & 0.234 & 0.384 & 0.372 & 0.213 & 0.362 & 0.331 & 0.184 & 0.347 & 0.305 & 0.165\\
                             & DIMVC   & 0.444 & 0.431 & 0.284 & 0.421 & 0.417 & 0.259 & 0.382 & 0.333 & 0.208 & 0.369 & 0.313 & 0.192 & 0.269 & 0.240 & 0.114 \\
                             & DCP   & 0.409 & 0.444 & 0.258 & 0.369 & 0.415 & 0.232 & 0.366 & 0.384 & 0.204 & 0.297 & 0.328 & 0.150 & 0.188 & 0.199 & 0.017 \\
                             & DSIMVC   & 0.298 & 0.300 & 0.240 & 0.306 & 0.298 & 0.251 & 0.284 & 0.273 & 0.223 & 0.270 & 0.255 & 0.216 & 0.267 & 0.244 & 0.218 \\
                             & DVIMC   & 0.467 & 0.464 & 0.301 & 0.451 & 0.434 & 0.284 & 0.418 & 0.390 & 0.253 & 0.381 & 0.333 & 0.210 & 0.271 & 0.239 & 0.214 \\
                             \rowcolor{lightgray}
                             & \textbf{SI$^3$}   & \textbf{0.494} & \textbf{0.487} & \textbf{0.324} & \textbf{0.483} & \textbf{0.462} & \textbf{0.312} & \textbf{0.445} & \textbf{0.410} & \textbf{0.266} & \textbf{0.426} & 0.381 & \textbf{0.244} & \textbf{0.398} & 0.334 & \textbf{0.222} \\ \hline
\multirow{8}{*}{\rotatebox{90}{\textbf{COIL100}}}       & PMIMC   & 0.712 & 0.890 & 0.650 & 0.691 & 0.877 & 0.632 & 0.671 & 0.869 & 0.615 & 0.626 & 0.846 & 0.560 & 0.641 & 0.848 & 0.575 \\
                             & CPSPAN  & 0.670 & 0.864 & 0.613 & 0.652 & 0.856 & 0.595 & 0.627 & 0.843 & 0.551 & 0.624 & 0.836 & 0.556 & 0.602 & 0.823 & 0.542\\
                             & GIMVC   & 0.734 & 0.904 & 0.655 & 0.711 & 0.897 & 0.610 & 0.709 & 0.896 & 0.617 & \textbf{0.714} & 0.897 & 0.632 & \textbf{0.689} & 0.879 & 0.585\\
                             & DIMVC   & 0.727 & 0.927 & 0.731 & 0.724 & 0.927 & 0.722 & 0.677 & 0.911 & 0.694 & 0.624 & 0.875 & 0.612 & 0.584 & 0.869 & 0.538\\
                             & DCP   & 0.604 & 0.837 & 0.542 & 0.572 & 0.820 & 0.485 & 0.533 & 0.789 & 0.424 & 0.448 & 0.735 & 0.320 & 0.358 & 0.651 & 0.219\\
                             & DSIMVC   & 0.581 & 0.801 & 0.505 & 0.514 & 0.758 & 0.430 & 0.498 & 0.749 & 0.412 & 0.487 & 0.742 & 0.401 & 0.483 & 0.716 & 0.402\\
                             & DVIMC   & 0.776 & 0.942 & 0.770 & \textbf{0.776} & \textbf{0.940} & \textbf{0.767} & 0.742 & 0.929 & 0.744 & 0.690 & 0.908 & \textbf{0.686 }& 0.625 & 0.878 & 0.606 \\
                             \rowcolor{lightgray}
                             & \textbf{SI$^3$}   & \textbf{0.786} & \textbf{0.944} & \textbf{0.774} & 0.766 & 0.938 & 0.764 & \textbf{0.757} & \textbf{0.933} & \textbf{0.756}  & 0.692 & \textbf{0.909} & 0.680 & 0.685 & \textbf{0.895} & \textbf{0.666}\\ \hline
\multirow{8}{*}{\rotatebox{90}{\textbf{Multi-Fashion}}}       & PMIMC   & 0.771 & 0.768 & 0.676 & 0.702 & 0.749 & 0.635 & 0.656 & 0.725 & 0.596 & 0.634 & 0.726 & 0.588 & 0.613 & 0.730 & 0.570 \\
                             & CPSPAN   & 0.652 & 0.737 & 0.602  & 0.661 & 0.721 & 0.587  & 0.637 & 0.713 & 0.580  & 0.554 & 0.658 & 0.493  & 0.574 & 0.639 & 0.482  \\
                             & GIMVC   & 0.709 & 0.765 & 0.625  & 0.740 & 0.757 & 0.635 & 0.680 & 0.718 & 0.572  & 0.672 & 0.660 & 0.536  & 0.568 & 0.569 & 0.421  \\
                             & DIMVC   & 0.707 & 0.754 & 0.627  & 0.734 & 0.766 & 0.658  & 0.734 & 0.761 & 0.656  & 0.671 & 0.666 & 0.569  & 0.634 & 0.661 & 0.556  \\
                             & DCP   & 0.795 & 0.837 & 0.727  & 0.794 & 0.808 & 0.709  & 0.747 & 0.769 & 0.648  & 0.667 & 0.670 & 0.556  & 0.552 & 0.558 & 0.299  \\
                             & DSIMVC   & \textbf{0.886} & 0.859 & 0.786  & 0.844 & 0.826 & 0.745  & 0.788 & 0.775 & 0.688  & 0.750 & 0.731 & 0.651  & 0.653 & 0.657 & 0.553  \\
                             & DVIMC   & 0.815 & 0.864 & 0.772  & \textbf{0.855} & \textbf{0.863} & \textbf{0.797}  & 0.822 & 0.837 & 0.757  & 0.779 &\textbf{ 0.806} & 0.706  & 0.752 & 0.763 & 0.663  \\
                             \rowcolor{lightgray}
                             & \textbf{SI$^3$}   & \textbf{0.886} & \textbf{0.881} & \textbf{0.831} & 0.854 & 0.860 & 0.794 & \textbf{0.840} & \textbf{0.841} & \textbf{0.772} & \textbf{0.783} & 0.805 & \textbf{0.708} & \textbf{0.779} & \textbf{0.764} & \textbf{0.679} \\ \hline
\end{tabular}
\label{tab：performance}
\end{table*}

TABLE~\ref{tab：performance} summarizes clustering performance across four benchmark datasets under varying missing rates. Several key observations highlight the motivation and novelty of our approach:
\begin{itemize}
    \item \textbf{Effectiveness of selective imputation.} Our method consistently surpasses both imputation-based and imputation-free baselines, validating that judiciously imputing only where sufficient evidence exists not only recovers missing associations but also avoids the noise typically introduced by indiscriminate imputation.  
    \item \textbf{Robustness under severe missingness.} While imputation-free methods (e.g., DVIMC) exhibit drastic degradation when missing rates are high (e.g., ACC on Caltech7-5V drops from 0.872 to 0.516 as the missing rate increases from 0.1 to 0.5), our method maintains stable performance, demonstrating that the proposed informativeness-based imputation position selection strategy is especially valuable when available evidence becomes sparse. Detailed visualized analysis follows in Q2.
    \item \textbf{Efficiency and scalability.} Compared to methods that cautiously integrate imputation (e.g., DSIMVC), our approach achieves higher clustering accuracy with relative lower computational overhead. This shows that our framework not only improves reliability but also scales better to larger or more incomplete datasets. 
\end{itemize}
To further examine the robustness of different methods under increasing missing rates, we plot the clustering performance (accuracy) as bar charts based on the results in TABLE~\ref{tab：performance}. Among the baselines, PMIMC is the leading imputation-based method, while DVIMC often delivers competitive or superior results among imputation-free approaches.
\begin{figure*}[t]
    \centering
    \subfloat[Caltech7-5V\label{fig:barcaltech}]{
        \includegraphics[width=0.47\columnwidth]{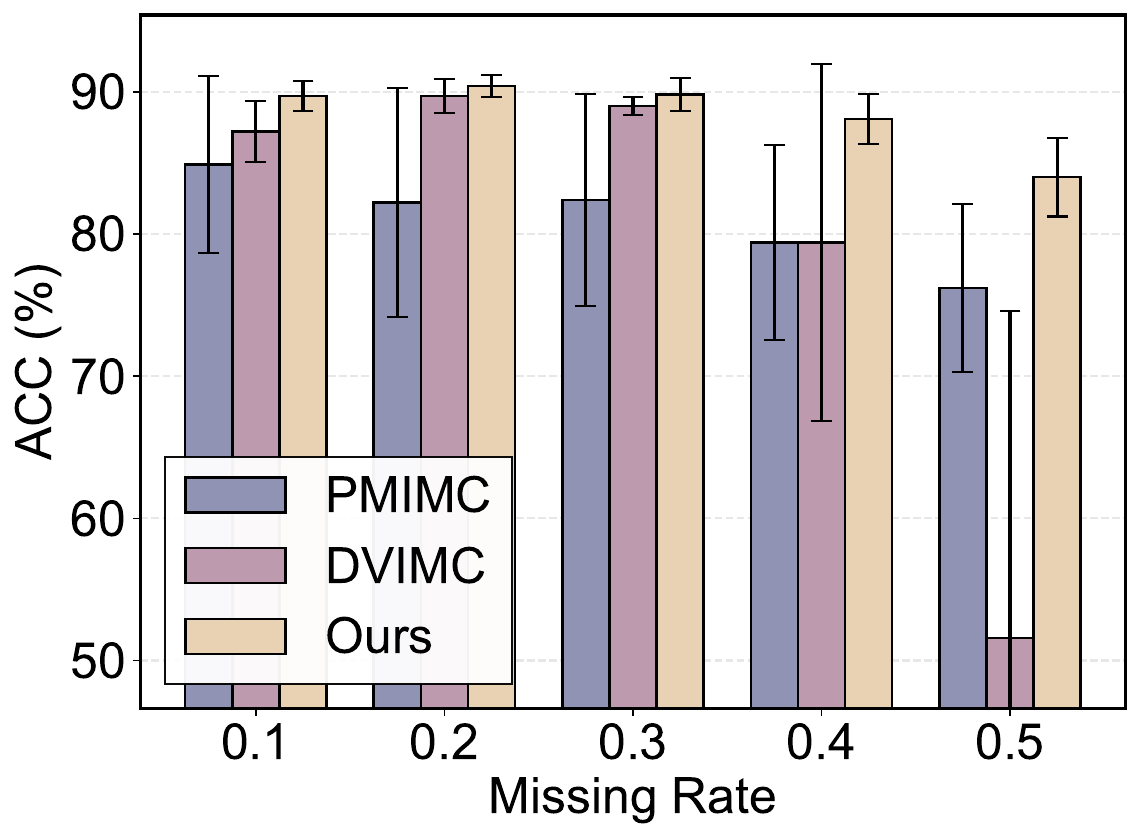}
    }
    \hfill
    \subfloat[Scene-15\label{fig:barscene}]{
        \includegraphics[width=0.47\columnwidth]{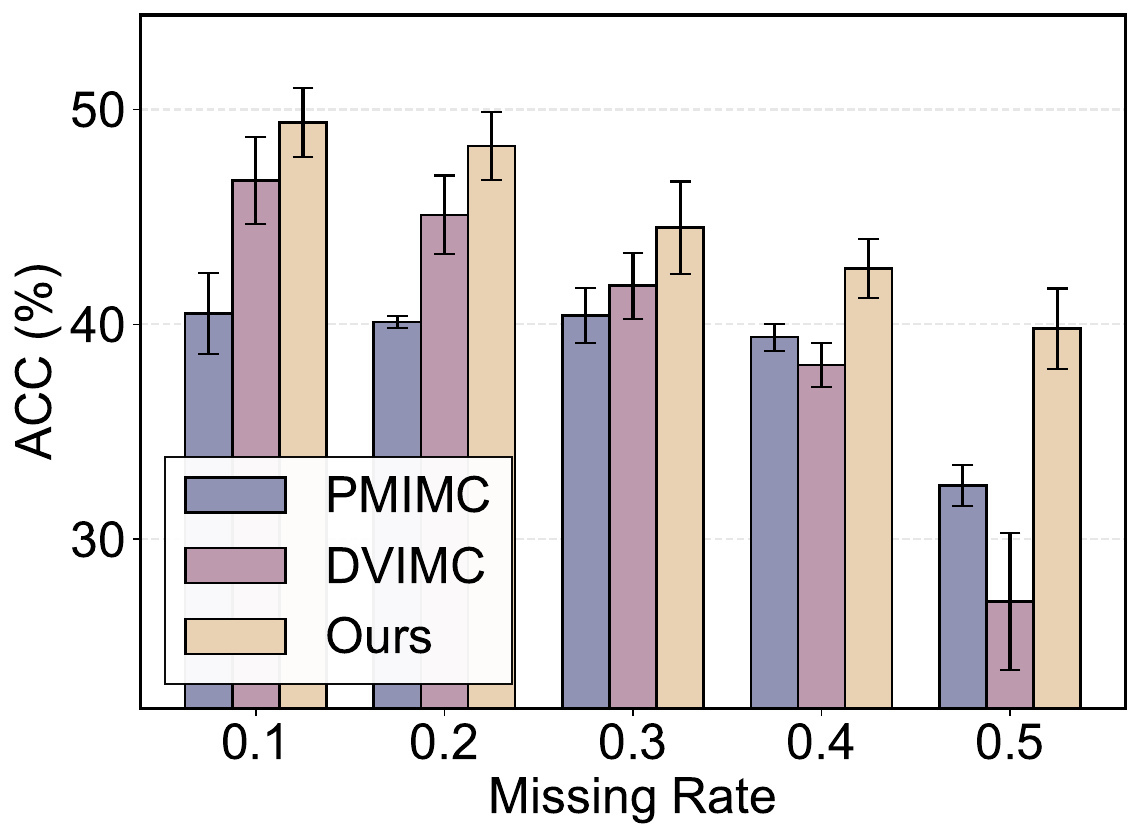}
    }
    \hfill
    \subfloat[COIL100\label{fig:barcoil}]{
        \includegraphics[width=0.47\columnwidth]{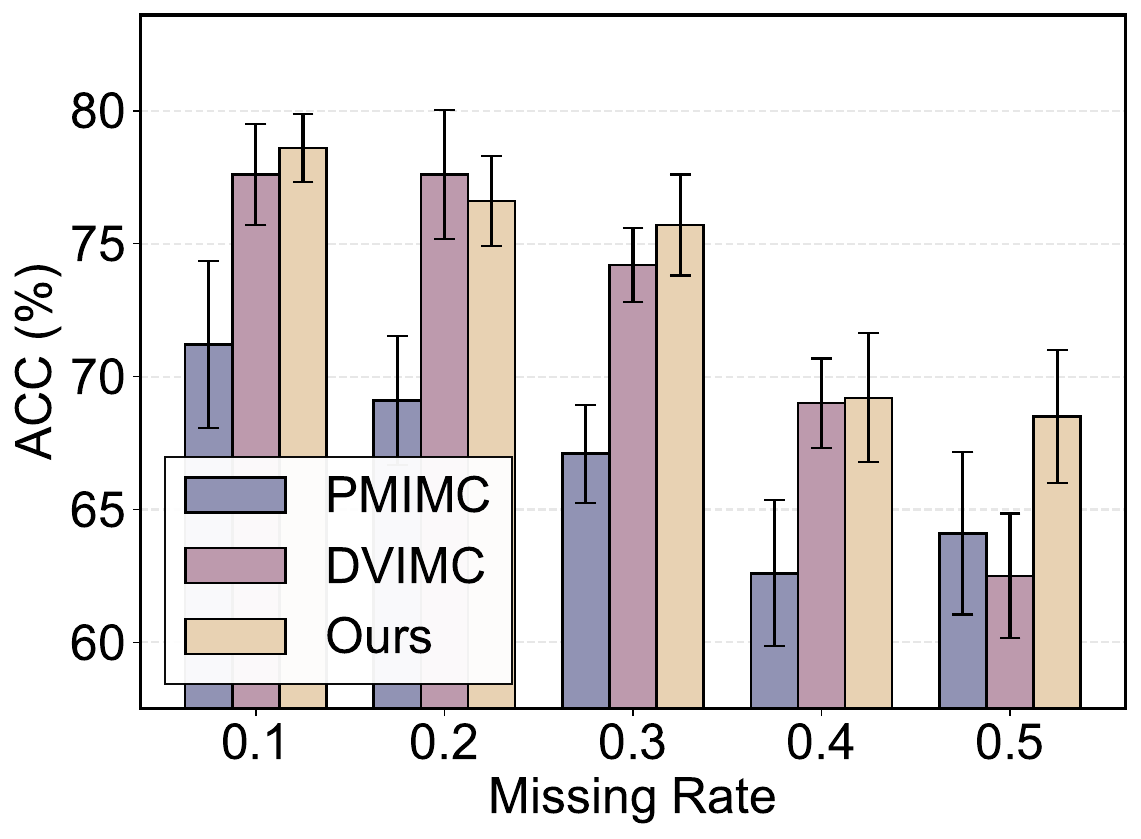}
    }
    \hfill
    \subfloat[Multi-Fashion\label{fig:barfashion}]{
        \includegraphics[width=0.47\columnwidth]{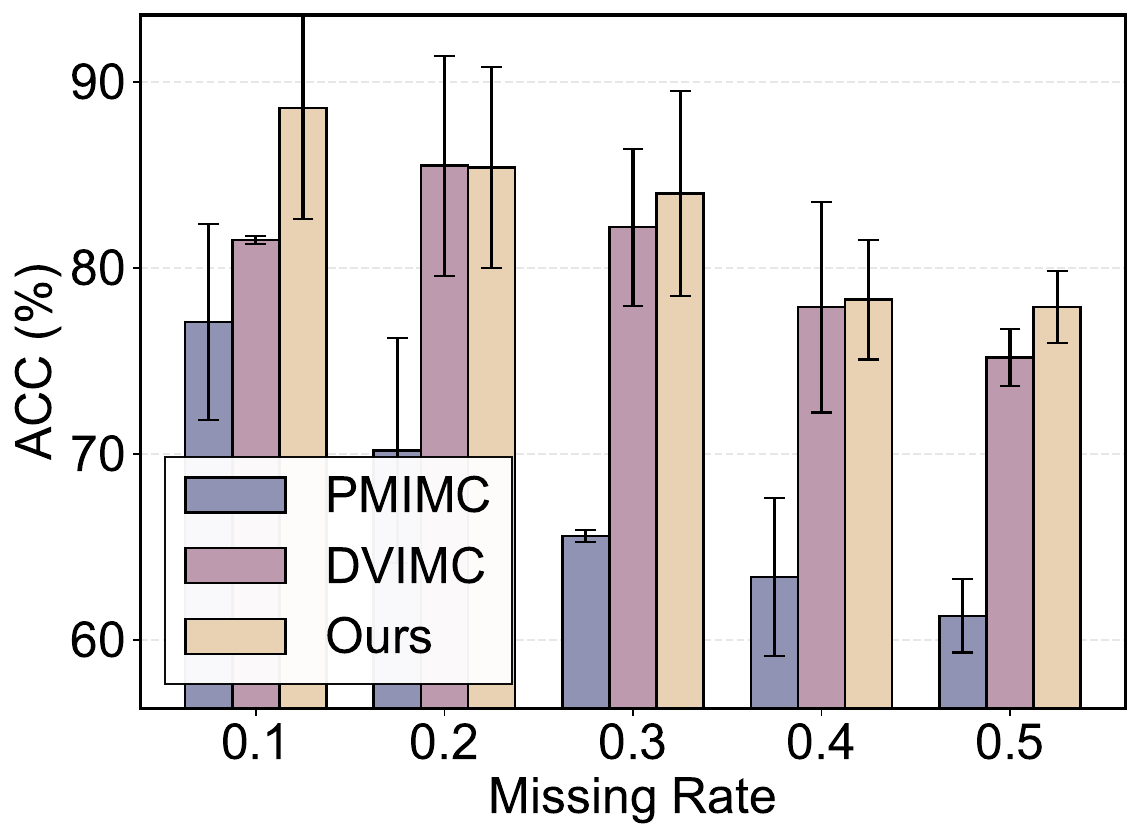}
    }
    \caption{Bar charts of accuracy under different missing rates (0.1 to 0.5) on four datasets, PMIMC and DVIMC are selected as representative imputation-based and imputation-free methods, respectively, both generally exhibiting competitive or near–state-of-the-art performance.}
    \label{fig:performance_visual}
\end{figure*}

As shown in Fig.~\ref{fig:performance_visual}, imputation-based methods like PMIMC deliver suboptimal performance across low missing rates due to unavoidable bias introduced by indiscriminate imputation; however, their accuracy remains relatively stable at higher missing levels. In contrast, imputation-free methods such as DVIMC achieve competitive performance at low missing rates, but exhibit sharp drops as missing rates increase. Our method consistently maintains the highest accuracy bars across all datasets and missing rates, showing only minimal reduction even at 0.5 missing rate, which demonstrates superior robustness without sacrificing performance under severe incompleteness.

\paragraph{Q2: Informativeness and Interpretability}
\begin{figure*}[t]
    \centering
    \subfloat[Heatmap of position-level informativeness scores\label{fig:info_heatmap}]{
        \includegraphics[width=0.46\textwidth]{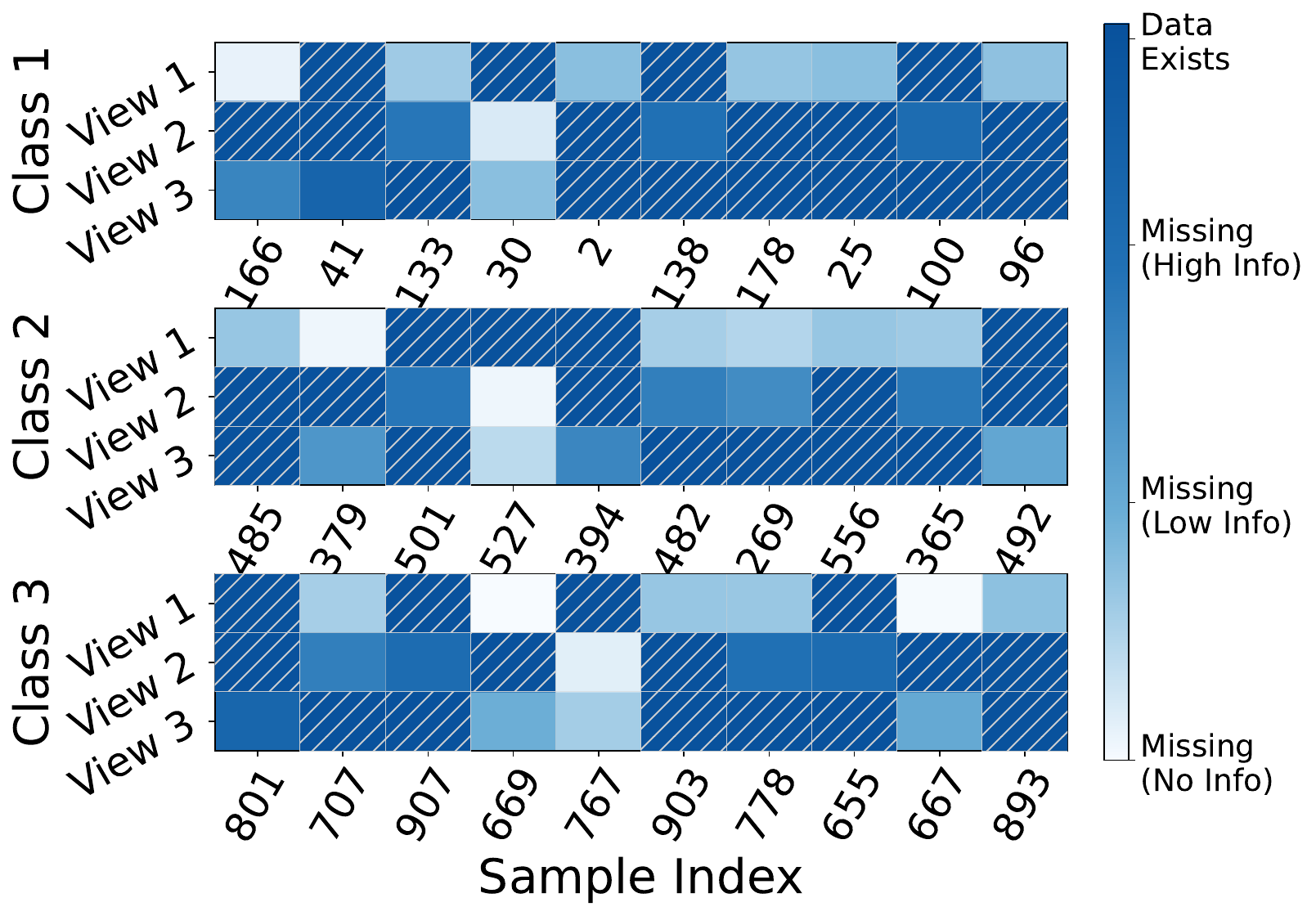}
    }\hfill
    \subfloat[Distributions under different missing rates\label{fig:info_violin}]{
        \includegraphics[width=0.46\textwidth]{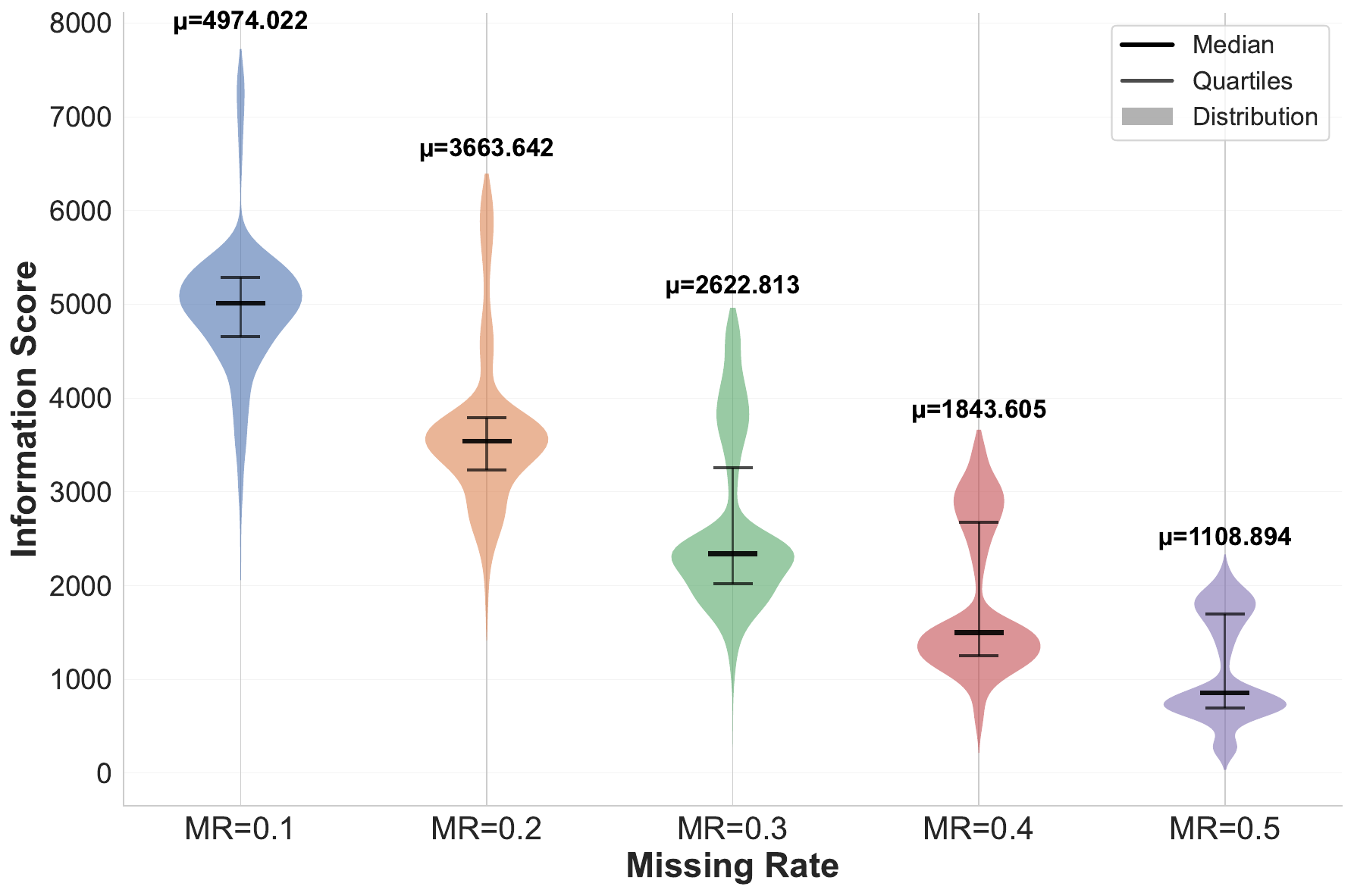}
    }
    \caption{Visualization of informativeness score on Scene-15. 
    (a) Heatmap of position-level scores with $\eta=0.5$, where deeper blue indicates higher score. Cells with diagonal hatching correspond to observed position, which are excluded from computation.
    (b) Violin plots of informativeness score distributions across different missing rates. 
    As missing rate increases, the score distribution shifts toward lower values and becomes more polarized.}
    \label{fig:info_visualization}
\end{figure*}

To better understand how informativeness score guide selective imputation, we provide both position-level and distribution-level visualizations on the Scene-15 dataset (Fig.~\ref{fig:info_visualization}). These analyses jointly reveal the heterogeneous informativeness of missing positions and its evolution with varying missing rates, offering direct evidence for our design motivation.

Fig.~\ref{fig:info_heatmap} presents the heatmap of informativeness scores under a 0.5 missing rate, where each cell denotes the score of a missing position. 
Darker shades correspond to higher score, while the darkest cells with diagonal hatching indicate observed data that do not require imputation. 
The missing pattern is configured such that view 1 suffers from the highest probability of missingness, followed by view 2 and view 3.
This visualization provides a fine-grained perspective on how the availability of intra-view neighbors and cross-view correspondences influences the imputation-relevant informativeness of each missing position.

In detail, the heatmap demonstrates clear heterogeneity across views: missing positions in view 2 or view 3 often achieve higher scores (e.g., positions (138, 2) and (801, 3)), owing to the availability of more intra-view neighbors and stronger cross-view support.
In contrast, samples missing in view 1 typically exhibit lower informativeness score, as the heavy missingness reduces the size of the support set. These observations suggest that samples missing in view 1 with limited cross-view evidence are inherently less suitable for imputation, highlighting the necessity of an Informativeness-based Imputation position selection mechanism rather than indiscriminate filling.

Beyond individual cases, we further examine how informativeness scores are distributed across different missing rates. Figure~\ref{fig:info_violin} shows violin plots of the score distributions. To account for the increasing number of missing positions at higher missing rates, the distributions are plotted based on proportions rather than absolute counts, facilitating a fair comparison across varying levels of incompleteness.

Several trends are noteworthy in the violin plots. First, as the missing rate increases from 0.1 to 0.5, the mean informativeness score drops sharply (from approximately 4974 to 1108), and the overall distribution shifts downward, reflecting the diminishing quantity and reliability of available support.
Second, at low missing rates ($\eta$ = 0.1, 0.2), the scores concentrate into a single sharp peak at high values, indicating that most missing positions can be confidently estimated.
As the missingness grows ($\eta$ = 0.3, 0.4, 0.5), the distribution gradually flattens and splits into two modes—one corresponding to well-supported positions and the other to poorly-supported ones—with the lower mode becoming dominant at $\eta$ = 0.5.
This transition from a unimodal to a pronounced bimodal pattern reveals the increasing heterogeneity in imputation reliability and highlights the necessity of selectively imputing positions with sufficiently high informativeness score.

The above findings provide strong empirical justification for our method. The position-level heatmap demonstrates that imputation-relevant informativeness is highly uneven across different views and positions, while the violin plots reveal that such unevenness grows into polarization as missingness increases. Together, these analyses confirm that imputation should not be uniformly applied. Instead, it is crucial to prioritize positions with abundant and reliable evidence while avoiding those in the low-score regime.   

\paragraph{Q3: Robustness and Imputation Ratio Sensitivity}

\begin{figure}[t]
    \centering
    \subfloat[Caltech7-5V ($\eta=0.1$)\label{fig:a}]{
        \includegraphics[width=0.45\columnwidth]{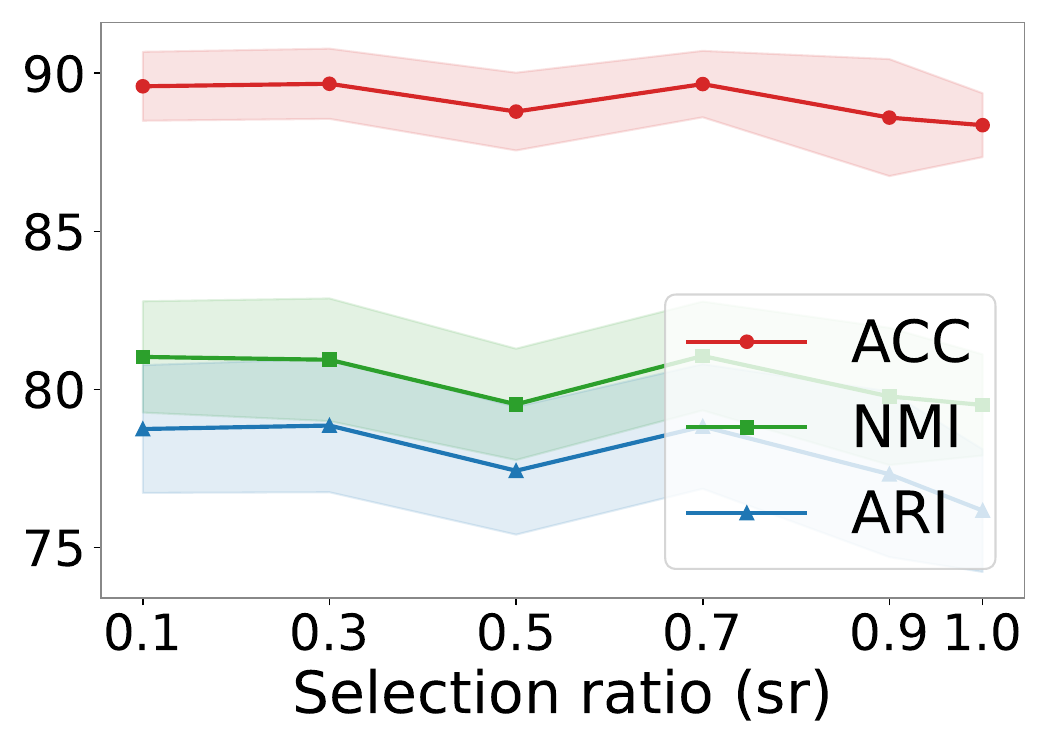}
    }
    \hfill
    \subfloat[Caltech7-5V ($\eta=0.5$)\label{fig:b}]{
        \includegraphics[width=0.45\columnwidth]{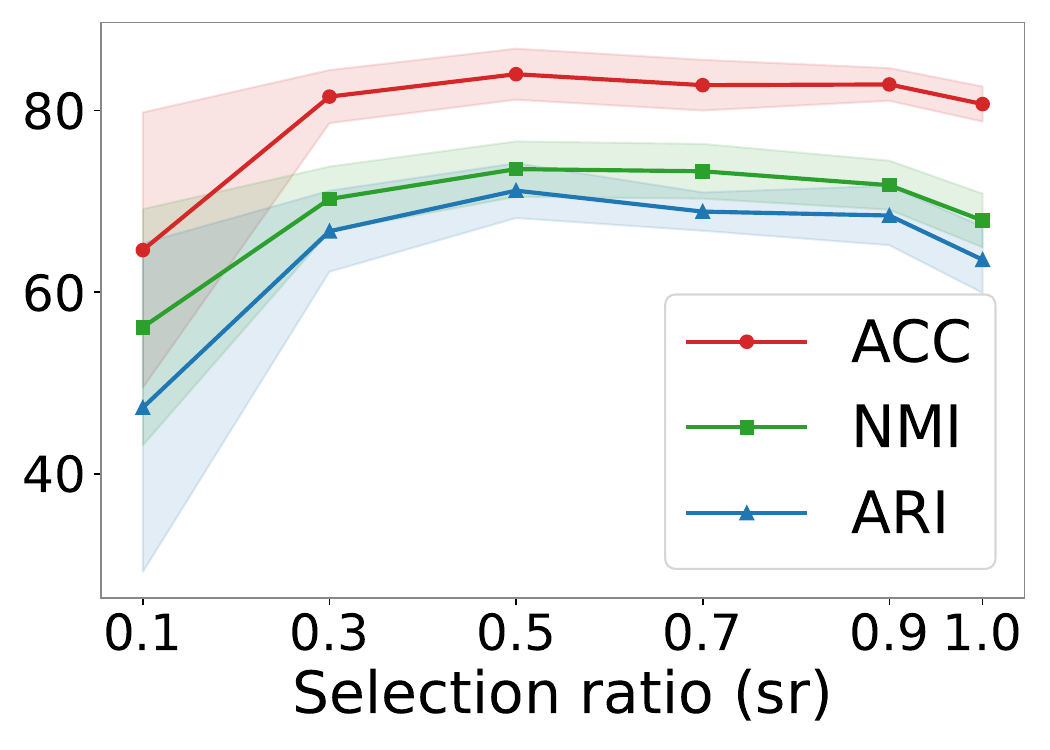}
    }
    \\
    \subfloat[Scene-15 ($\eta=0.1$)\label{fig:c}]{
        \includegraphics[width=0.45\columnwidth]{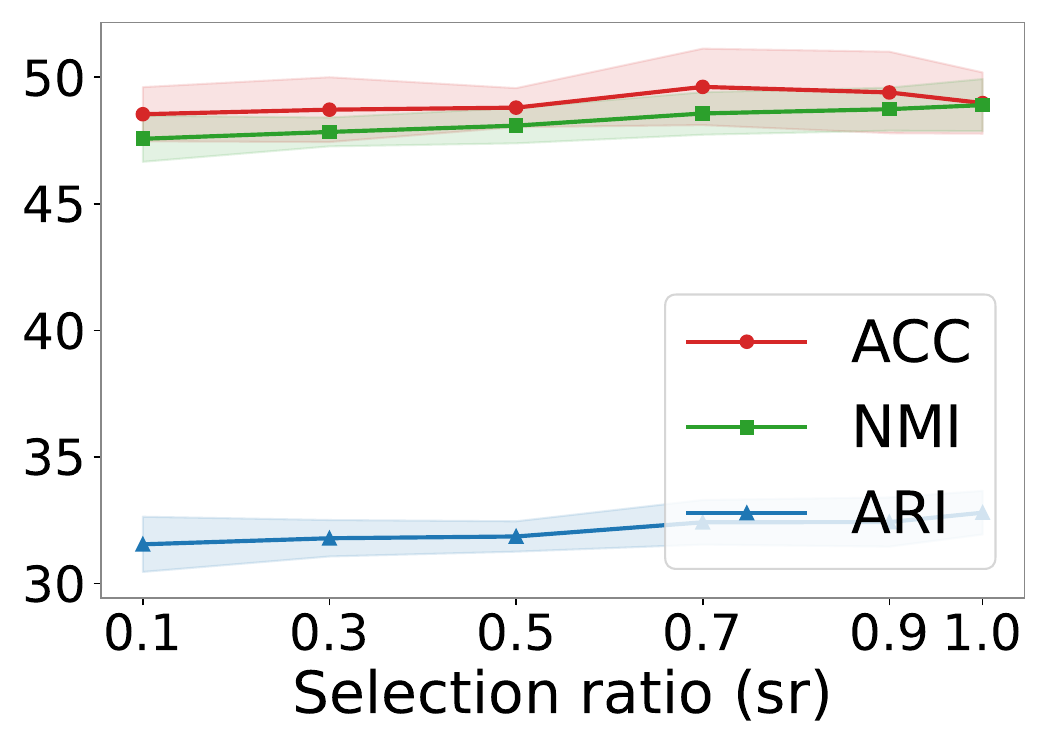}
    }
    \hfill
    \subfloat[Scene-15 ($\eta=0.5$)\label{fig:d}]{
        \includegraphics[width=0.45\columnwidth]{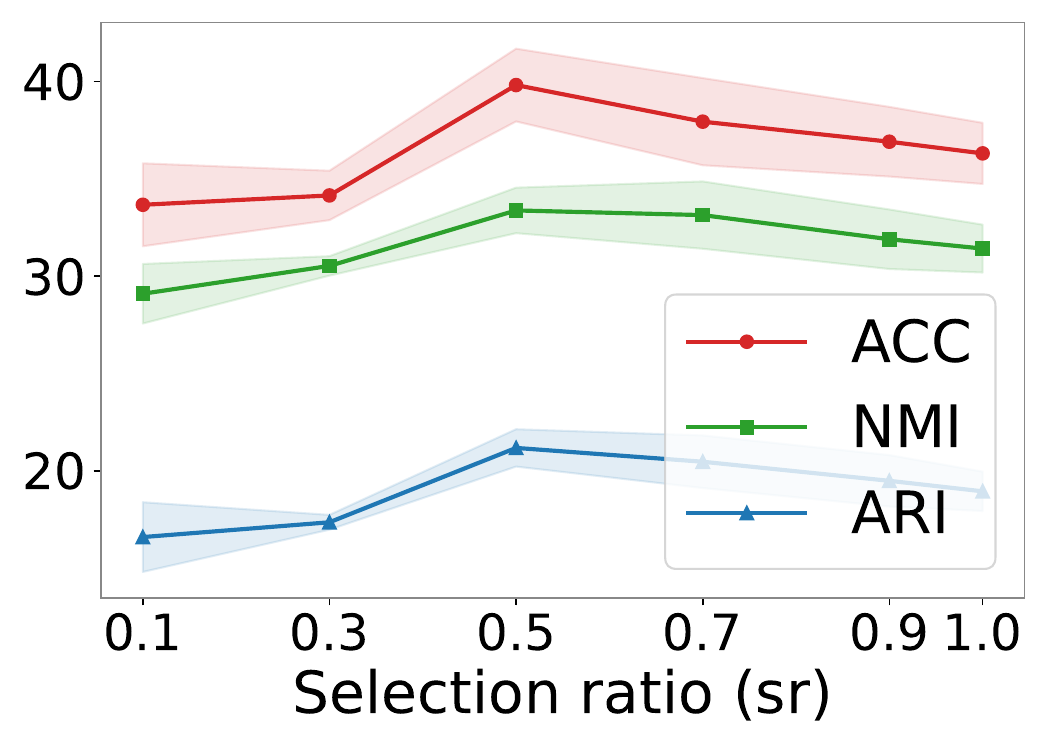}
    }
    \caption{Selection ration experiment on two datasets under low (0.1) and high (0.5) missing rates.}
    \label{fig:select_exp}
\end{figure}

To further examine the effect of selective imputation, we study how varying the proportion of imputed positions influences clustering performance. Specifically, we adjust the selection ratio by changing the threshold of informativeness scores, and conduct experiments on Caltech7-5V and Scene-15 under both low ($\eta=0.1$) and high ($\eta=0.5$) missing rates. The results are presented in Fig.~\ref{fig:select_exp}.  

Several consistent patterns can be observed. Under low missing rates (Fig.~\ref{fig:a} and \ref{fig:c}), performance remains relatively stable across a wide range of selection ratios. This suggests that when sufficient observations are available, the model benefits from strong supervision from observed data, and the potential noise introduced by imputation has only a marginal influence. In this case, even a high proportion of imputations does not substantially harm clustering quality.  

When the missing rate becomes higher ($\eta=0.5$), the role of selective imputation becomes more critical. On Caltech7-5V (Fig.~\ref{fig:b}), performance is still comparatively stable even at large selection ratios. This robustness can be attributed to the dataset’s rich multi-view structure (with 5 views), where complementary views provide reliable cues that reduce the negative impact of inaccurate imputations. In contrast, Scene-15 (Fig.~\ref{fig:d}) exhibits a clear rise-then-fall pattern: performance initially improves as more missing positions are filled, but starts to degrade once the ratio exceeds a certain threshold. This behavior indicates a trade-off: moderate imputation compensates for the structural collapse caused by severe missingness, whereas excessive imputation introduces accumulated noise that offsets its benefits.  

\paragraph{Q4: Generality and Plug-in Capability of IBSI}
\begin{figure}[t]
    \centering
    \subfloat[GIMVC+IBSI\label{fig:pluga}]{
        \includegraphics[width=0.45\columnwidth]{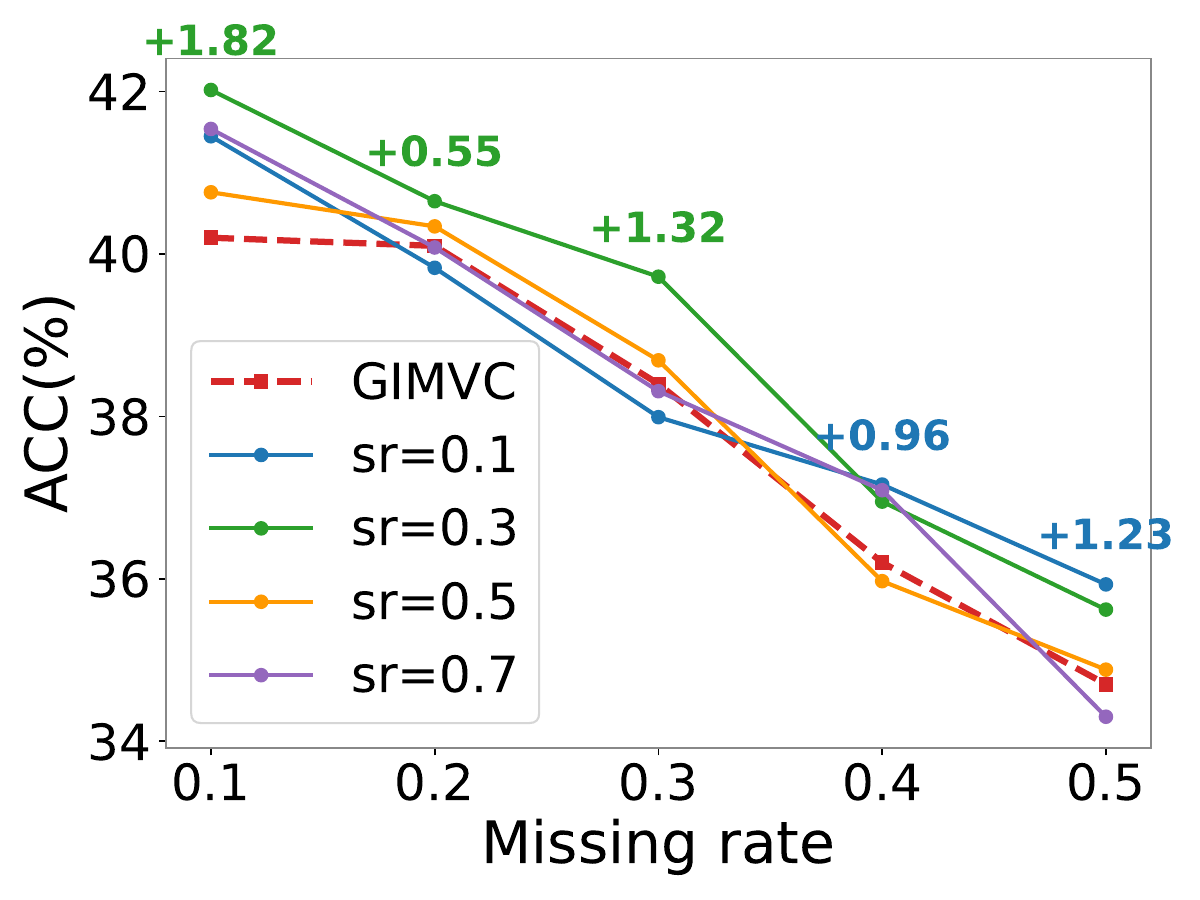}
    }
    \hfill
    \subfloat[DCP+IBSI\label{fig:plugb}]{
        \includegraphics[width=0.45\columnwidth]{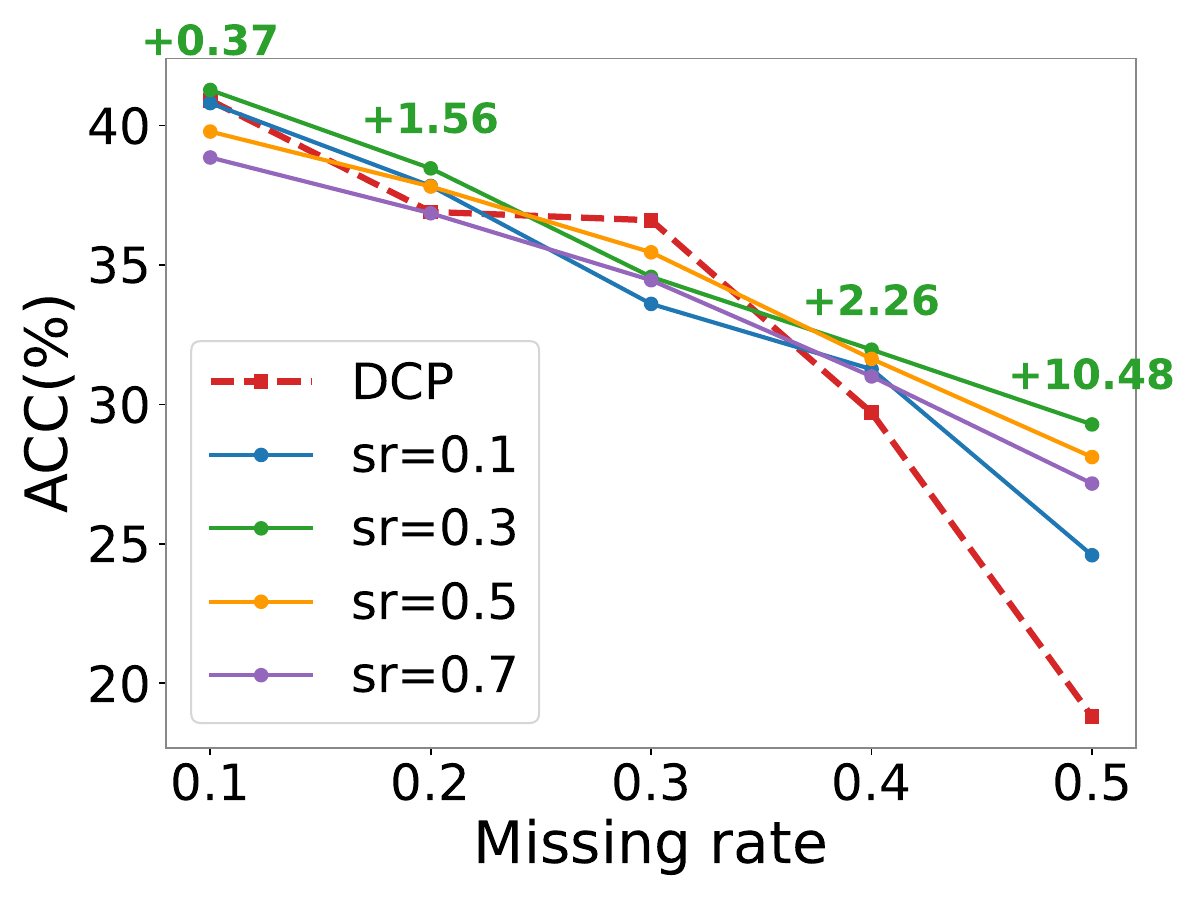}
    }
    \caption{Accuracy improvement of IBSI as a plug-in module on Scene-15. 
IBSI is integrated into (a) GIMVC (imputation-free) and (b) DCP (complete-sample-only). 
Numbers indicate the absolute ACC gains over each baseline at the optimal selection ratio (sr).}
    \label{fig:plug-in}
\end{figure}

\begin{figure}[t]
    \centering
    \subfloat[GIMVC+IBSI\label{fig:pluga_time}]{
        \includegraphics[width=0.45\columnwidth]{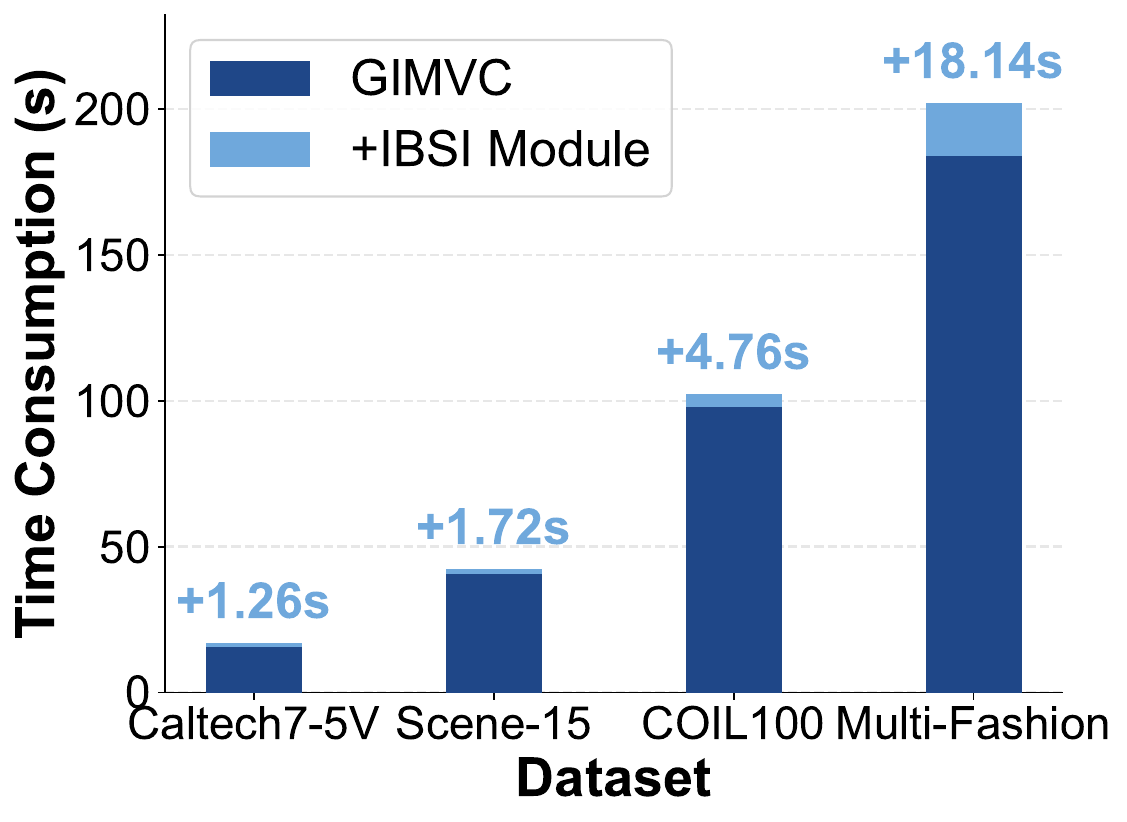}
    }
    \hfill
    \subfloat[DCP+IBSI\label{fig:plugb_time}]{
        \includegraphics[width=0.45\columnwidth]{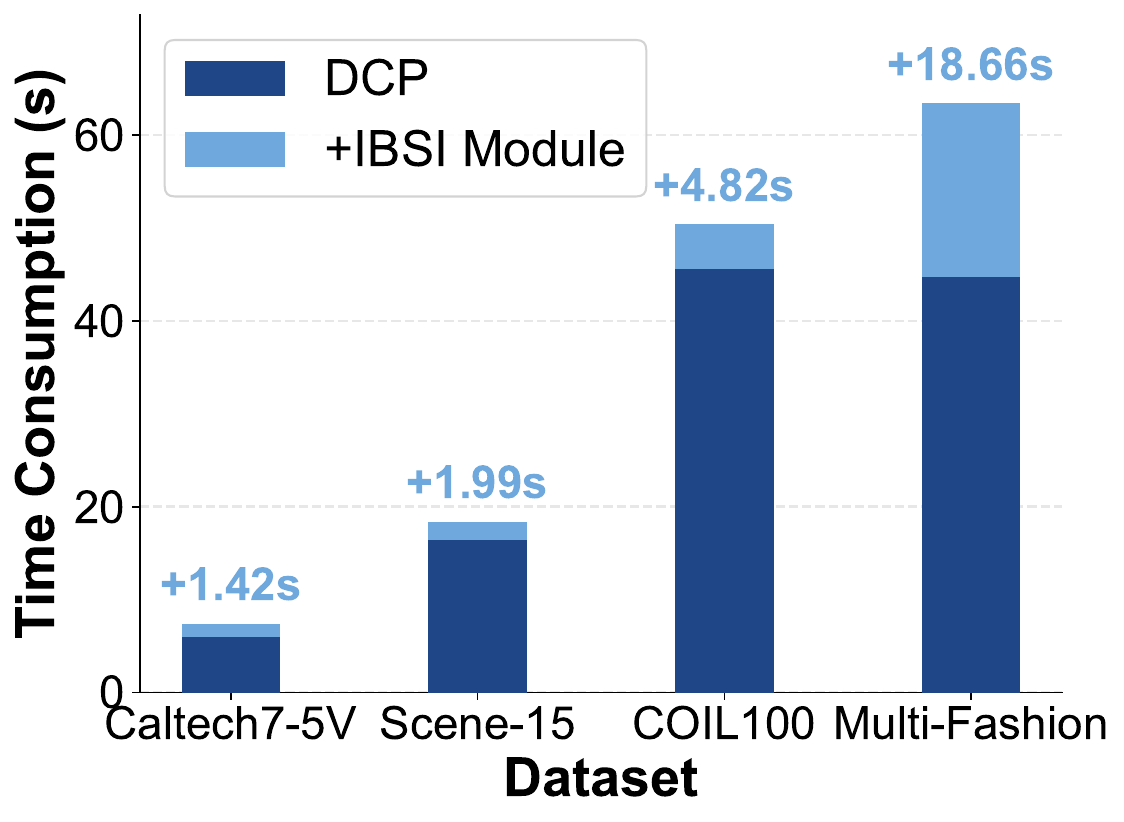}
    }
    \caption{Computational overhead of IBSI as a plug-in module across four datasets. 
Numbers indicate the absolute increase in running time over the baseline, 
evaluated under 0.3 selection ratio (sr) and $\eta=0.5$.}
    \label{fig:plug-in-time}
\end{figure}
In this section, we further evaluate the generality and plug-in potential of the proposed Informativeness-Based Selection and Indication (IBSI) module by integrating it into two representative frameworks with distinct philosophies: the imputation-free GIMVC and the complete-sample-only DCP. For fairness, both baselines use the same simple cross-view neighbor mean strategy in the raw feature space, ensuring that IBSI is the only additional component. 

As shown in Fig.~\ref{fig:plug-in}, IBSI consistently improves clustering accuracy across different missing rates on Scene-15, even when paired with such a primitive and non-parametric imputation procedure. The gains become more pronounced as the missing rate increases, indicating that IBSI effectively prioritizes informative and reliable positions when observations are sparse. This helps alleviate the performance degradation typically encountered in highly incomplete scenarios. These results confirm that IBSI is architecture-agnostic and does not rely on sophisticated reconstruction components, making it an easily deployable enhancement for a wide range of multi-view methods.

A further observation is that conservative selection ratios—i.e., only imputing positions with sufficiently high informativeness score—yield significantly better performance than aggressive imputations. This reflects the risk of introducing noise through unreliable estimates and highlights the benefit of focusing on high-confidence positions. Owing to the simplicity of the plug-in imputation mechanism, the optimal ratio is around 0.3, in contrast to about 0.5 in SI$^3$ where the imputation process is more principled and distribution-aware. This suggests that the appropriate selection ratio should be adapted to the sophistication of the underlying imputation module: simple schemes favor conservative ratios, while more advanced models can safely leverage a higher proportion of missing positions.

Beyond accuracy, we examine the time cost introduced by IBSI when used as a plug-in module. As shown in Fig.~\ref{fig:plug-in-time}, IBSI brings only a marginal overhead on all datasets. On Caltech7-5V, Scene-15, and COIL100, IBSI contributes no more than 5 additional seconds. Notably, the increase is proportionally smaller when integrated into the more complex GIMVC, where the baseline runtime is already substantial, and relatively larger for DCP, whose baseline computation is minimal. Nevertheless, the \emph{absolute} cost remains consistently small and nearly identical across methods, demonstrating that IBSI can be seamlessly incorporated regardless of the backbone’s complexity.

On the largest and most high-dimensional dataset, Multi-Fashion, IBSI introduces only about 18 seconds of extra computation. This increase is expected because the dataset has a significantly higher feature dimension (768), and our current plug-in implementation performs imputation in the raw feature space. However, this overhead can be further reduced by adopting more advanced strategies—such as performing selective imputation in latent space rather than directly on raw features.

Overall, the plug-in experiments reveal that IBSI offers both strong generality and excellent practicality: it provides substantial performance gains with only negligible computational cost, and its benefits are preserved across architectures of varying complexity. These qualities make IBSI a lightweight yet powerful component that can be readily integrated into diverse multi-view learning pipelines.

\paragraph{Q5: Sensitivity to the Regularization Coefficient} 
\begin{figure}[t]
    \centering
    \subfloat[Caltech7-5V\label{fig:alphacaltech}]{
        \includegraphics[width=0.47\columnwidth]{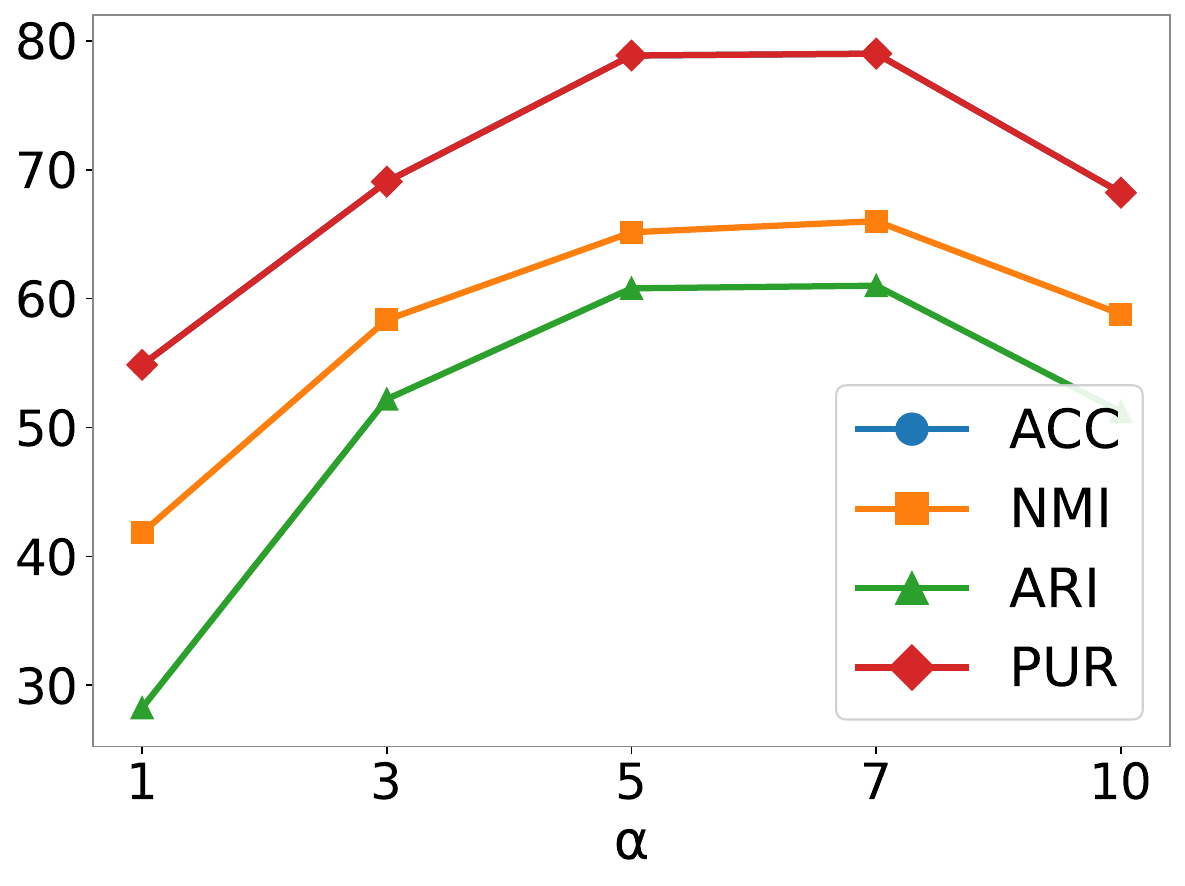}
    }
    \hfill
    \subfloat[Scene-15\label{fig:alphascene}]{
        \includegraphics[width=0.47\columnwidth]{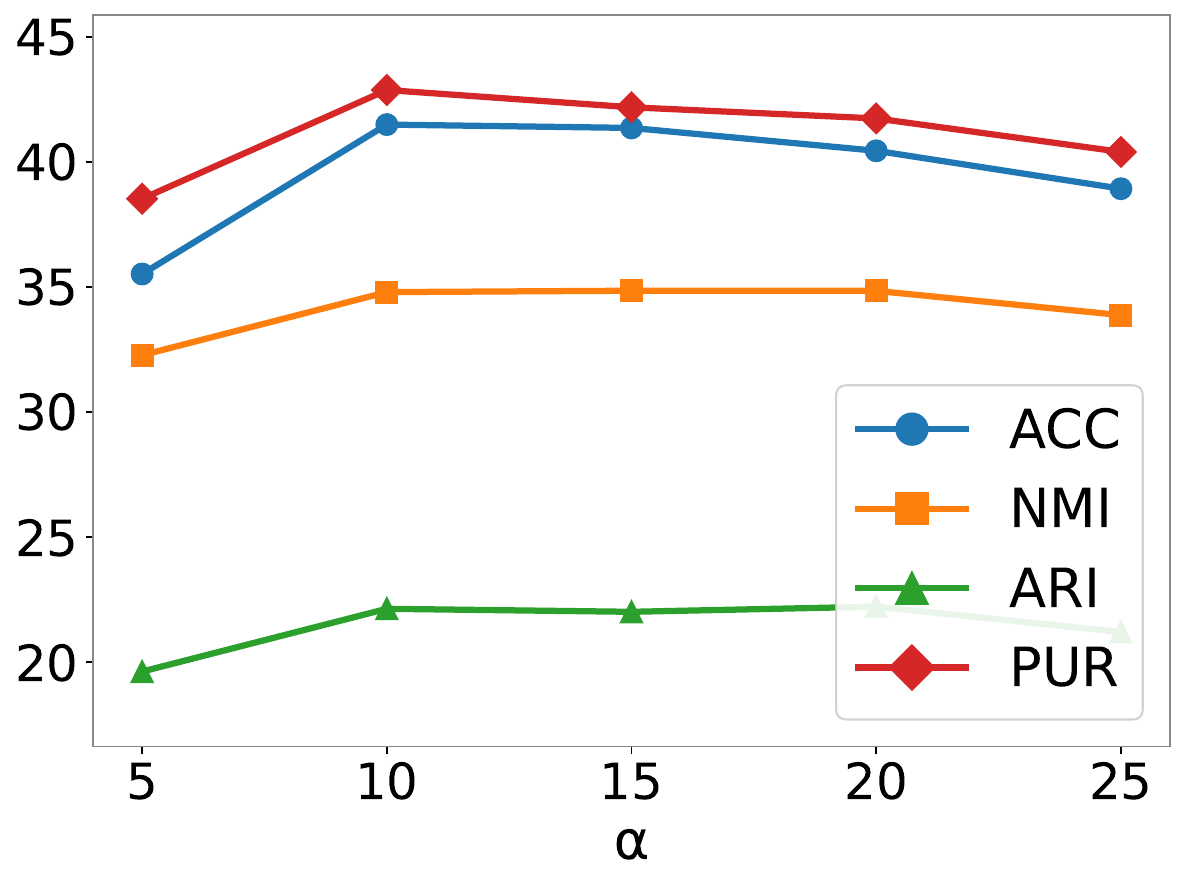}
    }
    \hfill
    \subfloat[COIL100\label{fig:alphacoil}]{
        \includegraphics[width=0.47\columnwidth]{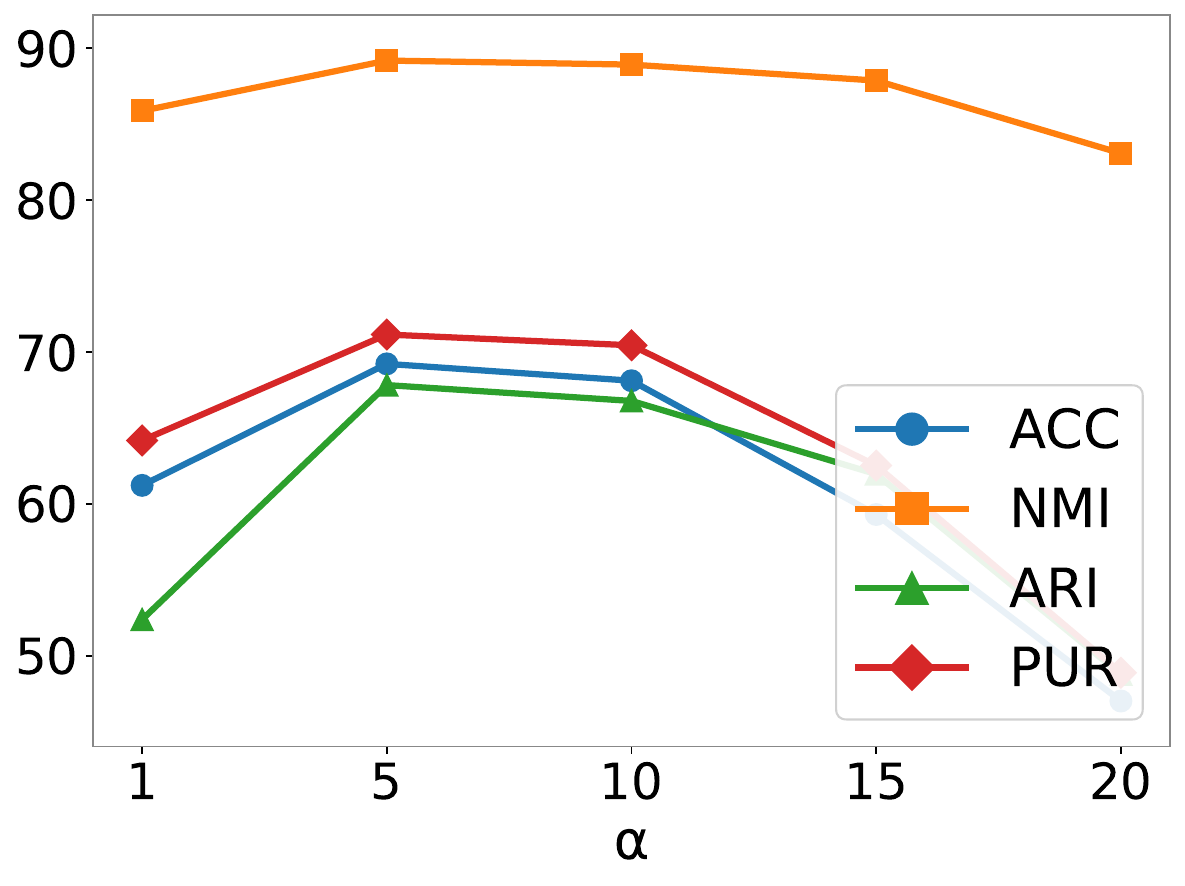}
    }
    \hfill
    \subfloat[Multi-Fashion\label{fig:alphafashion}]{
        \includegraphics[width=0.47\columnwidth]{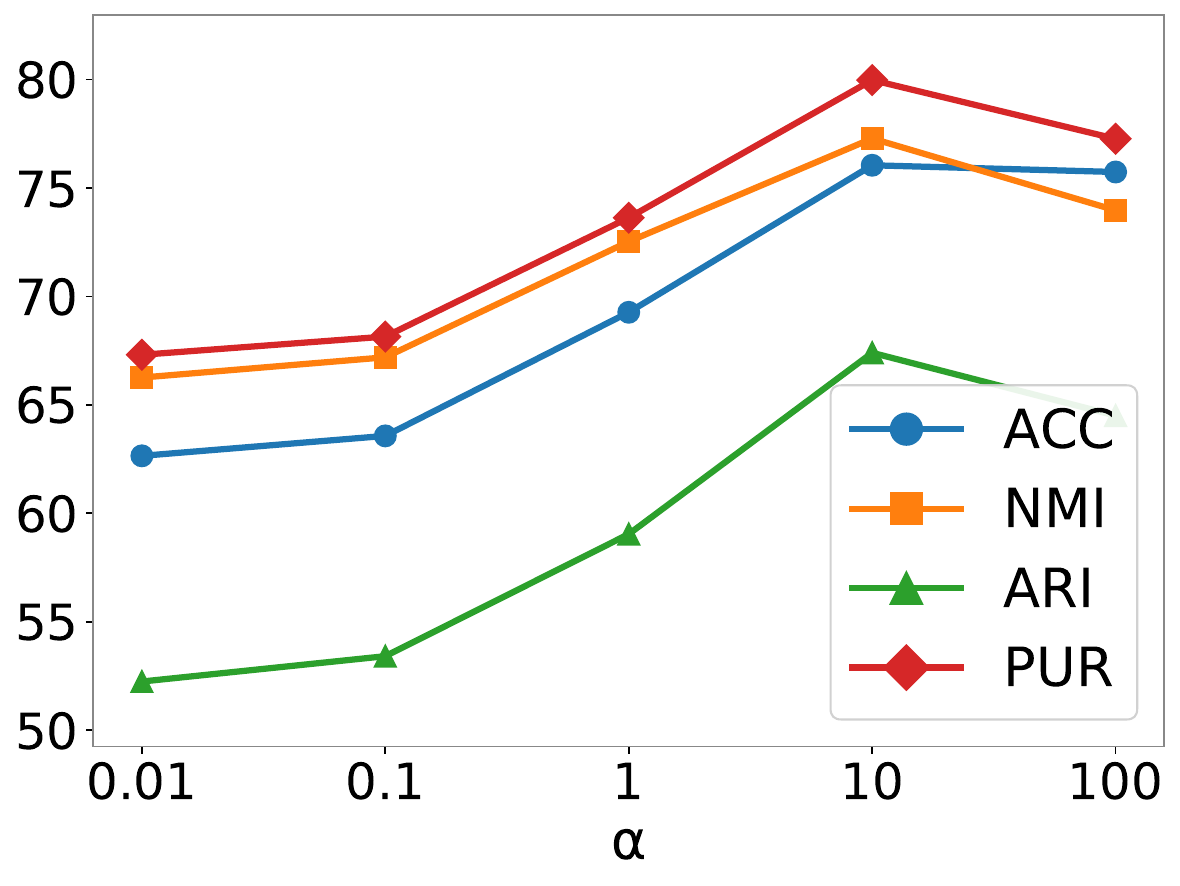}
    }
    \caption{Parameter sensitivity analysis of the balancing weight $\alpha$ between reconstruction and consistency across four datasets, conducted under a missing rate of 0.5 and a selection ratio of 0.5.}
    \label{fig:alphaexp}
\end{figure}
To examine the impact of the regularization balance on model performance, we conduct experiments under a missing rate of 0.5 by varying the coefficient $\alpha$, which controls the trade-off between reconstruction and consistency objectives. The results, shown in Fig.~\ref{fig:alphaexp}, indicate that the clustering performance is sensitive to this parameter.
When $\alpha$ is too small, the model tends to underemphasize the consistency constraint, leading to unstable alignment among views and insufficiently discriminative representations. Conversely, an excessively large $\alpha$ makes the optimization overly dominated by the coherence term, which weakens the model’s ability to preserve view-specific information.
In both extremes, the aggregated representation becomes less effective for clustering. A moderate value of $\alpha$ achieves a better equilibrium between view coherence and representational diversity, yielding superior results. Empirically, we find that setting $\alpha$ in the range of 5 to 10 provides consistently stable and competitive performance across datasets.

\section{Conclusion}

In this paper, we propose SI$^3$, a simple yet effective informativeness-based selective imputation method for incomplete multi-view clustering. By quantifying imputation-relevant informativeness at each missing position, SI$^3$ imputes only well-supported positions and models imputation uncertainty within a variational inference framework. This approach strikes a principled balance between leveraging imputation benefits and mitigating potential adverse effects. Additionally, the Informativeness-Based Selection and Indication module functions as a lightweight and plug-in component, enabling seamless integration with a wide range of clustering paradigms. Extensive experiments demonstrate its effectiveness and generality across diverse datasets and baselines.

\section*{Acknowledgments}
This research was supported by the National Natural Science Foundation of China (Grant Nos. 62133012, 62303366) and Natural Science Basic Research Program of Shaanxi under Grant No.2023-JC-QN-0648.



\bibliographystyle{IEEEtran}
\bibliography{paper_refer}

\begin{IEEEbiography}[{\includegraphics[width=1in,height=1.25in,clip,keepaspectratio]{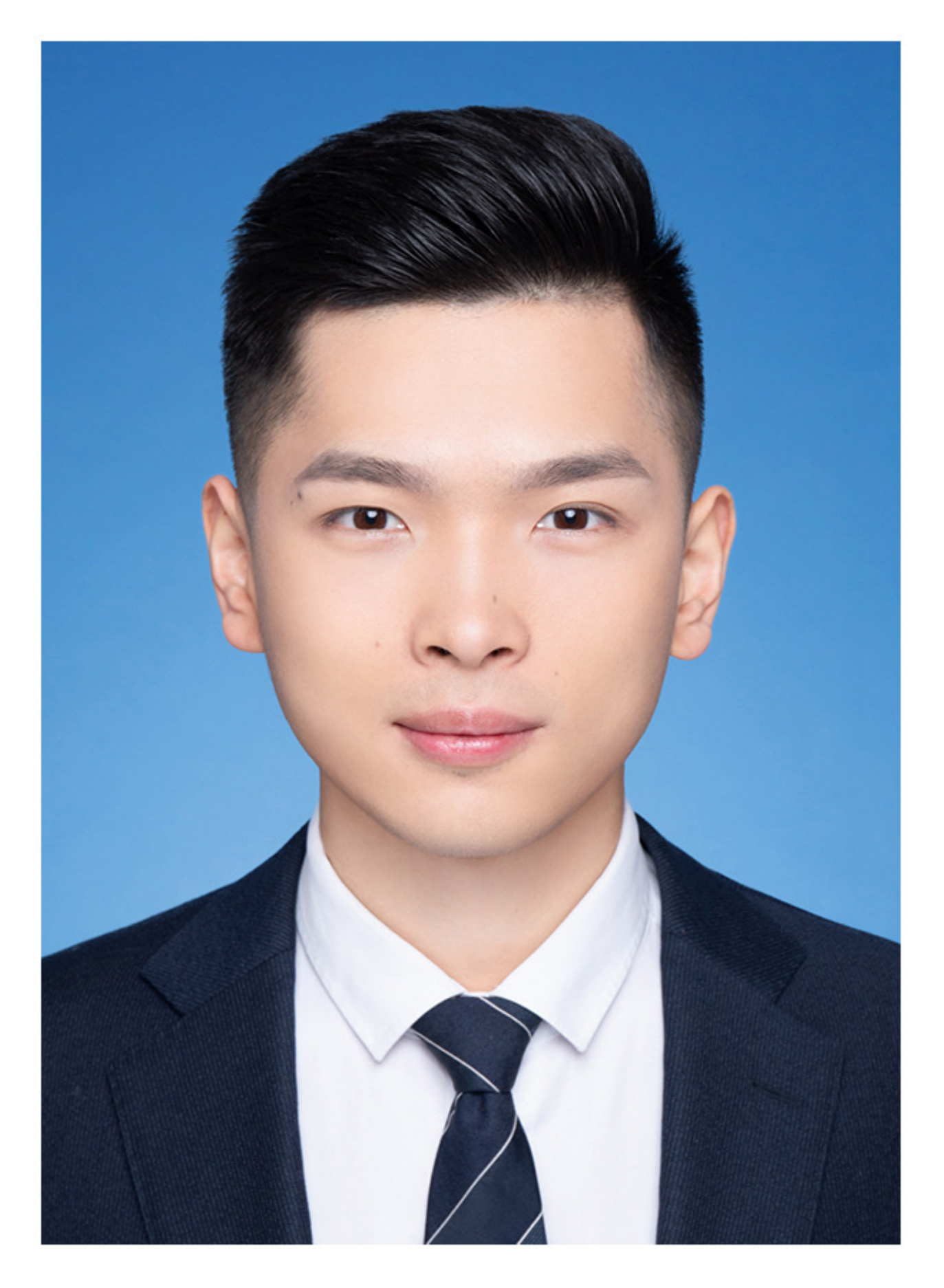}}]{Cai Xu}
    is currently an associate professor at the School of Computer Science and Technology, Xidian University. He received the Outstanding Paper Award at AAAI 24 as the first author. His research interests include trustworthy machine learning and multi-view learning.
\end{IEEEbiography}

\begin{IEEEbiography}[{\includegraphics[width=1in,height=1.25in,clip,keepaspectratio]{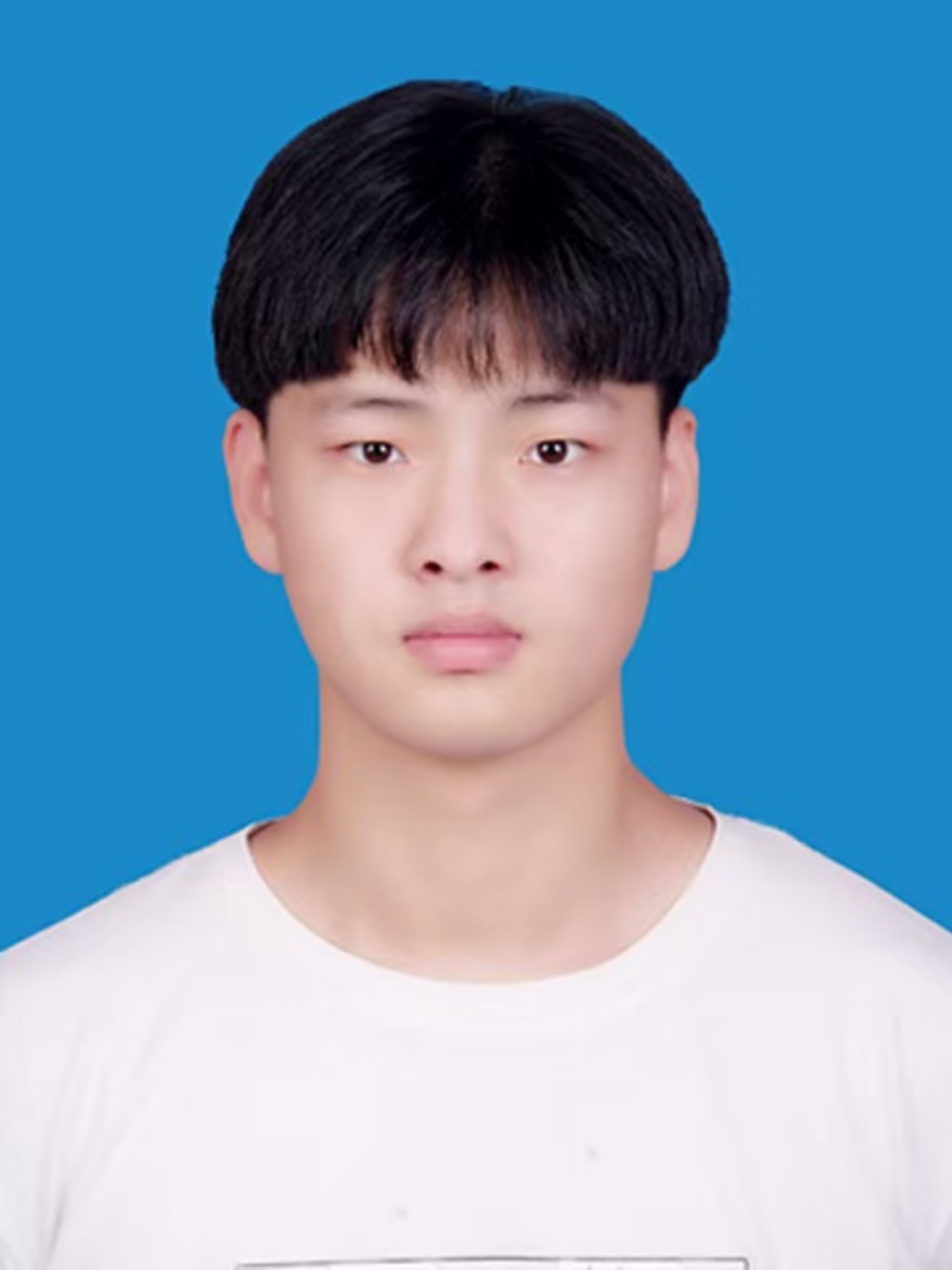}}]{Jinlong Liu}
    is currently a master’s student in Computer Science and Technology at Xidian University. His research interests focus on incomplete multi-view clustering and classification, with an emphasis on representation learning under missing and heterogeneous data.
\end{IEEEbiography}

\begin{IEEEbiography}[{\includegraphics[width=1in,height=1.25in,clip,keepaspectratio]{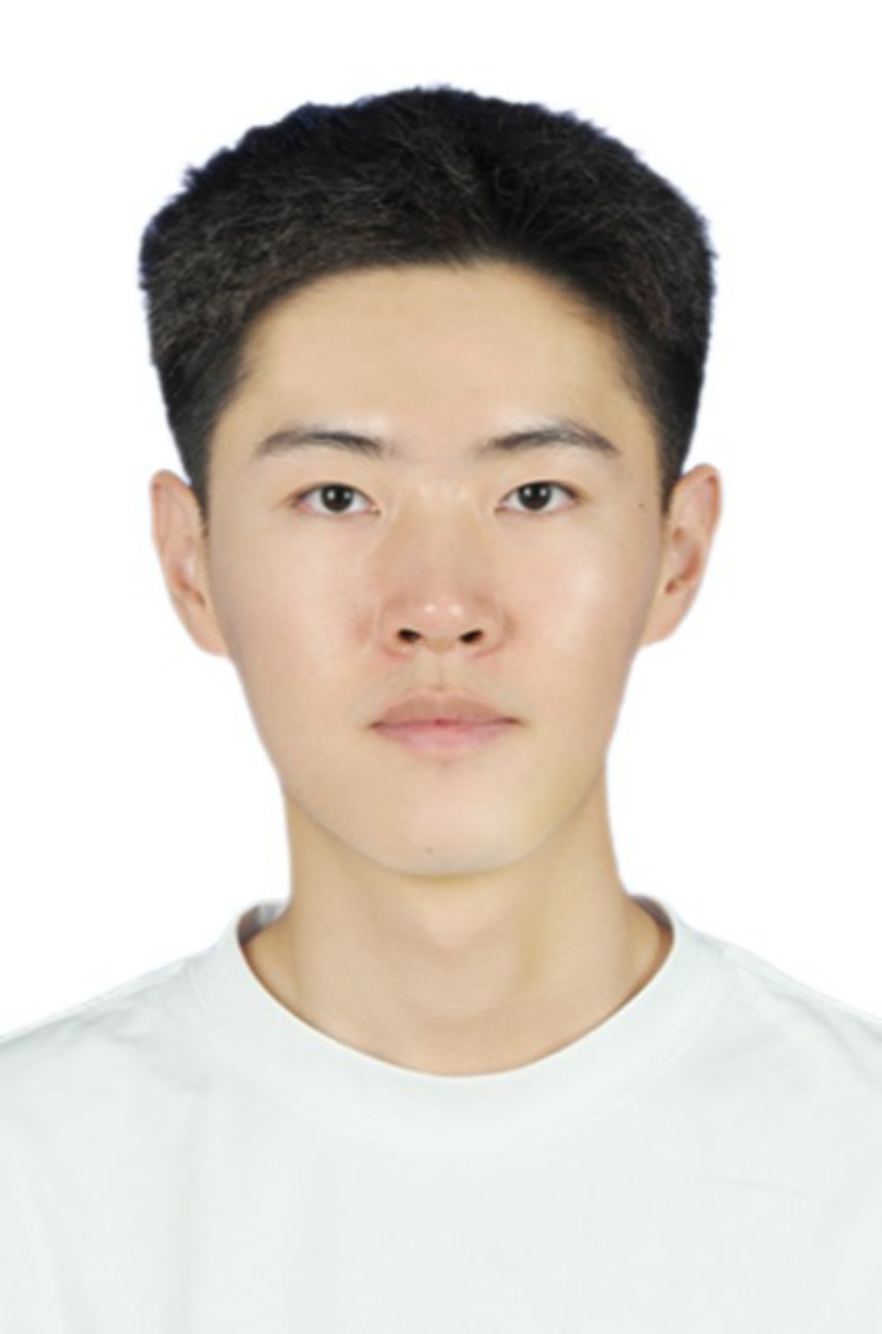}}]{Yilin Zhang}
	received the BS degrees in Computer Science and Technology from Xidian University, Xi'an, China, in 2023.   He is currently working toward the PhD degree with the School of Computer Science and Technology, Xidian University.   His research interests include multi-view learning, label-noise learning and uncertainty calibration.
\end{IEEEbiography}

\begin{IEEEbiography}[{\includegraphics[width=1in,height=1.25in,clip,keepaspectratio]{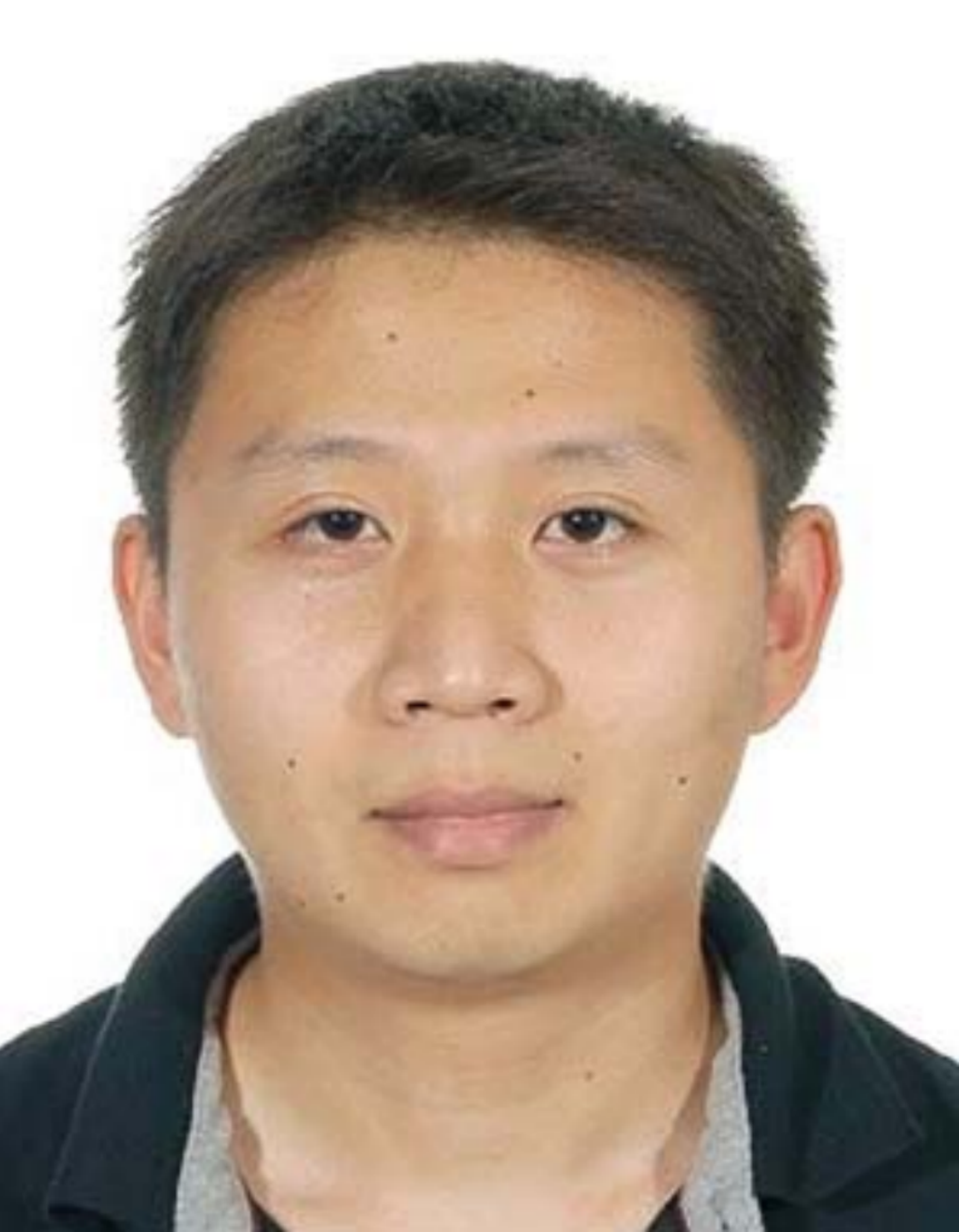}}]{Ziyu Guan} is currently a professor with the School of Computer Science and Technology, Xidian University. He is an Associate Editor for well-known journals such as IEEE TKDE, KAIS and JMLC. His research interests include attributed graph mining and search, machine learning, expertise modeling and retrieval, and recommender systems.
\end{IEEEbiography}

\begin{IEEEbiography}[{\includegraphics[width=1in,height=1.25in,clip,keepaspectratio]{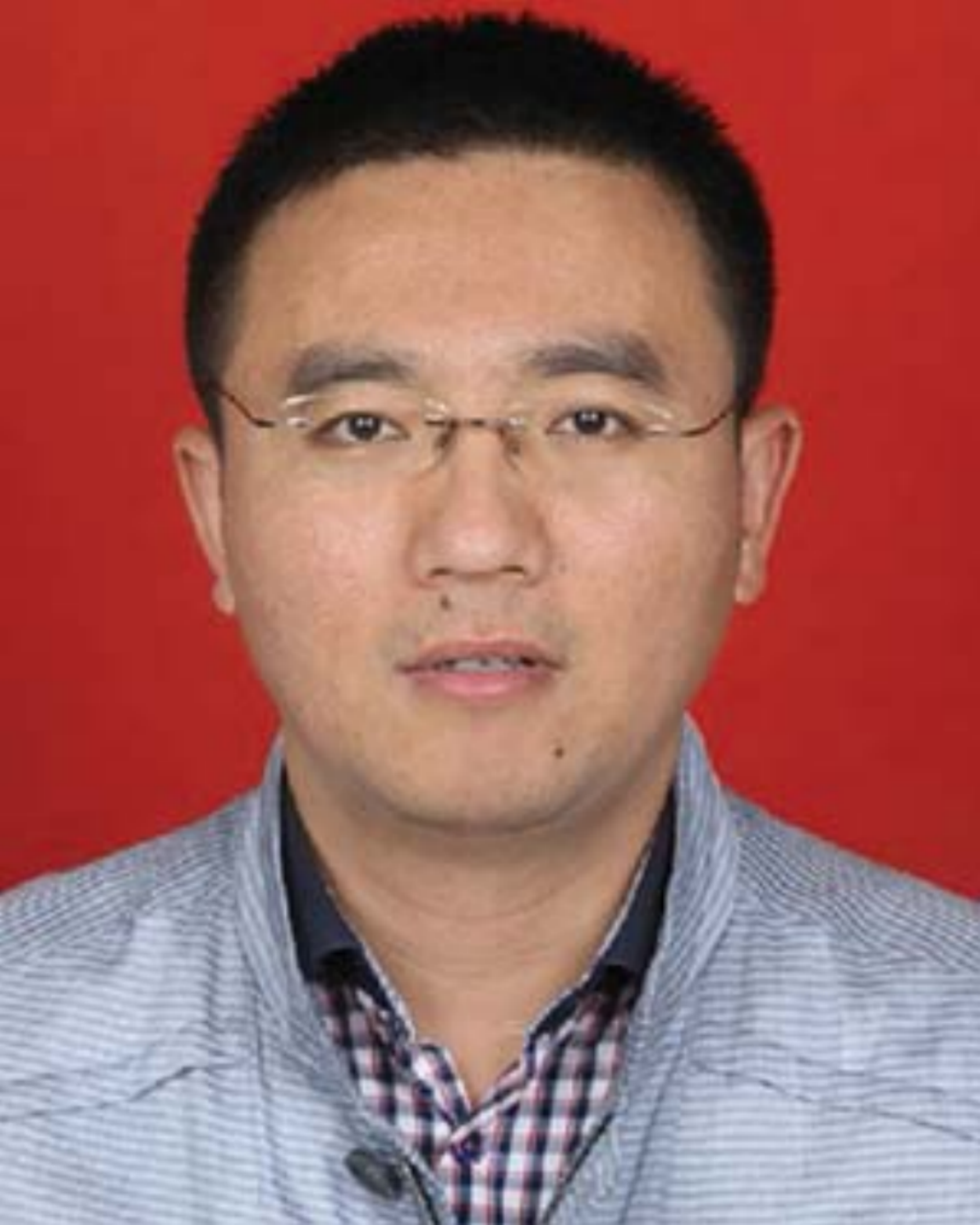}}]{Wei Zhao}
	is currently a professor with the School of Computer Science and Technology, Xidian University. He is an Associate Editor for well-known journals such as IEEE TKDE and TNNLS. His research interests include direction is pattern recognition and intelligent systems, with specific interests in attributed graph mining and search, machine learning, signal processing, and precision guiding technology.
\end{IEEEbiography}


\begin{IEEEbiography}[{\includegraphics[width=1in,height=1.25in,clip,keepaspectratio]{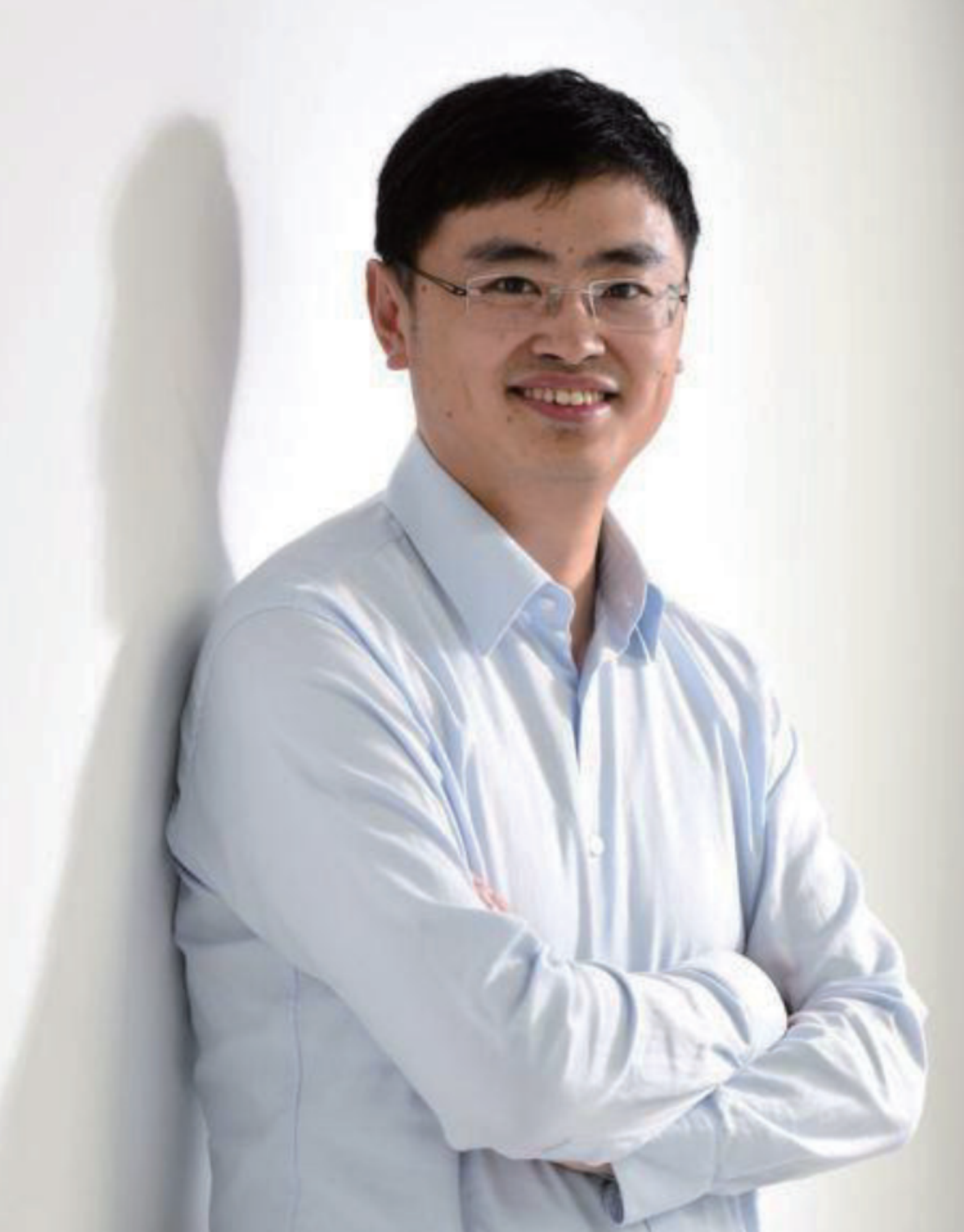}}]{Xiaofei He}
 is currently a Professor at the State Key Lab of CAD\&CG, Zhejiang University, and the CEO of FABU Technology Co., Ltd.  His research interests include machine learning, deep learning, and autonomous driving.  He has authored/co-authored more than 200 technical papers with over 45,000 times citations on Google Scholar.  He was awarded the Best Paper Award at AAAI 2012 and is a Fellow of IAPR.
\end{IEEEbiography}

\vfill

\end{document}